\pdfoutput=1
\documentclass[10pt,twocolumn,letterpaper]{article}

\usepackage[pagenumbers]{cvpr}

\usepackage[utf8]{inputenc}
\usepackage[T1]{fontenc}
\usepackage{graphicx}
\usepackage{booktabs}
\usepackage{multirow}
\usepackage{makecell}
\usepackage{colortbl}
\usepackage{amsmath,amssymb}
\usepackage{mathtools}
\usepackage{bm}
\usepackage{xcolor}
\usepackage{enumitem}
\usepackage{microtype}
\usepackage{xspace}
\usepackage{nicefrac}
\usepackage{siunitx}
\usepackage[many]{tcolorbox}
\usepackage{algorithm}
\usepackage[noend]{algpseudocode}

\usepackage{cuted}
\usepackage{caption}
\makeatletter
\g@addto@macro\normalsize{%
  \setlength\abovedisplayskip{5pt plus 2pt minus 2pt}%
  \setlength\belowdisplayskip{5pt plus 2pt minus 2pt}%
  \setlength\abovedisplayshortskip{0pt plus 2pt}%
  \setlength\belowdisplayshortskip{3pt plus 2pt minus 2pt}%
}
\makeatother

\definecolor{cvprblue}{rgb}{0.21,0.49,0.74}
\usepackage[pagebackref,breaklinks,colorlinks,allcolors=cvprblue]{hyperref}
\usepackage{placeins}

\newcommand{\method}{\textup{\textsc{Relay}}\xspace}

\definecolor{firebrick}{rgb}{0.698,0.133,0.133}
\definecolor{LightCyan}{HTML}{CCEEFF}   %
\definecolor{LightGray}{HTML}{EEEEEE}
\definecolor{tolgreen}{HTML}{228833}    %
\definecolor{tolblue}{HTML}{4477AA}     %
\definecolor{tolred}{HTML}{EE6677}      %
\definecolor{paleyellow}{HTML}{EEEEBB}
\definecolor{darkyellow}{HTML}{666633}
\definecolor{palegreen}{HTML}{CCDDAA}
\definecolor{darkgreen}{HTML}{225522}
\definecolor{paleblue}{HTML}{BBCCEE}
\definecolor{darkblue}{HTML}{222255}

\newcommand{\limref}[1]{\hyperref[lim:#1]{\textbf{\textcolor{firebrick}{(#1)}}}}

\newcommand{\iou}{\mathrm{IoU}}
\newcommand{\cosine}{\operatorname{cos}}

\definecolor{promptgray}{HTML}{CCCCCC}
\protected\def\promptspacing{\frenchspacing\spaceskip=\fontdimen2\font\xspaceskip=\fontdimen2\font\relax}
\newtcolorbox{promptbox}[2][]{
  breakable, enhanced, colback=#2!25!white, colframe=#2!80!black, boxrule=0.5pt, arc=2pt,
  left=5pt, right=5pt, top=3pt, bottom=3pt, fonttitle=\bfseries\small, coltitle=black,
  colbacktitle=#2!60!white, title={#1}, halign=left,
  fontupper=\footnotesize\ttfamily\promptspacing}
\newcommand{\promptrole}[1]{\par\addvspace{3pt}{\normalfont\bfseries\spaceskip=0pt\xspaceskip=0pt #1}\par\nobreak\addvspace{1pt}}
\newcommand{\promptfield}[1]{\textcolor{tolblue}{\{#1\}}}
\newcommand{\promptnote}[1]{{\normalfont\itshape\spaceskip=0pt\xspaceskip=0pt\color{gray}[#1]}}
\newcommand{\promptitem}{\par\hangindent=2\fontdimen2\font\hangafter=1 }
\newtcolorbox{promptfloat}[2][]{
  enhanced, colback=#2!25!white, colframe=#2!80!black, boxrule=0.5pt, arc=2pt,
  left=5pt, right=5pt, top=3pt, bottom=3pt, fonttitle=\bfseries\small, coltitle=black,
  colbacktitle=#2!60!white, title={#1}, fontupper=\footnotesize\ttfamily}

\AtBeginDocument{%
  \crefname{algorithm}{Algorithm}{Algorithms}
  \Crefname{algorithm}{Algorithm}{Algorithms}
}

\makeatletter
\def\cvprsect#1{\cvprsection[#1]{\texorpdfstring{\hskip -1em.~}{}#1}}
\makeatother

\def\confName{CVPR}
\def\confYear{2027}

\title{Towards Agentic Video Corpus Moment Retrieval}

\author{
Chenwei Xu\textsuperscript{1}\thanks{Correspondence to \texttt{cxu@u.northwestern.edu}.}\quad
Shang Wu\textsuperscript{1}\quad
Zhenyu Pan\textsuperscript{1}\quad
Jianshu Zhang\textsuperscript{1}\quad
Pranav Maneriker\textsuperscript{2}\\
Lie Lu\textsuperscript{2}\quad
Fan Du\textsuperscript{2}\quad
Manling Li\textsuperscript{1}\quad
Han Liu\textsuperscript{1}\\[0.4em]
\textsuperscript{1}Northwestern University\quad
\textsuperscript{2}Dolby
}

\begin{document}
\maketitle

\begin{abstract}
Video corpus moment retrieval (VCMR) requires finding the video and time interval that match a text query in a corpus of videos.
We present \method, a training-free framework for VCMR that shares memory across queries.
\method processes a batch of queries in three stages: search with a query bank, ranking with video notes, and localization with video prompt caching.
In the search stage, an embedding model scores each video by its best-matching frame.
The query bank penalizes videos that score highly for many other queries in the batch.
In the ranking stage, a vision-language model (VLM) writes notes on each candidate video's scenes, people, objects, and actions once.
Then, a large language model (LLM) reranks the candidates from these notes.
In the localization stage, a VLM predicts the time interval of the moment in the top-ranked video.
When multiple queries select the same video, prompt caching lets the VLM reuse its frames.
On the three VERIFIED benchmarks, \method outperforms the trained baselines in video retrieval R@1 without training.
Without any labeled moment, its predicted intervals are shorter than the annotated ones.
In VCMR, it achieves similar performance to VideoSearch-R1 on one benchmark and trails it at IoU 0.5 on the other two.
Human annotators also mark shorter moments than the original annotations of the benchmarks.
A single margin, calibrated on a few labeled queries, widens the predicted interval to the annotation convention of the benchmarks.
With it, \method outperforms the trained baselines in VCMR at IoU 0.5.
\vspace{-0.1in}

\end{abstract}

\section{Introduction}
\label{sec:intro}
\begin{figure}[t]
\centering
\includegraphics[width=\linewidth]{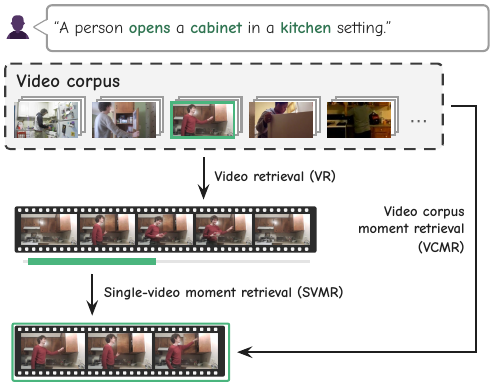}
\vspace{-0.22in}
\caption{\textbf{Video corpus moment retrieval (VCMR).}
Given a text query and a corpus of videos, VCMR retrieves the video that contains the moment the query describes.
It also localizes the start and end times of this moment.
VCMR joins video retrieval (VR), which finds the video, and single-video moment retrieval (SVMR), which finds the moment in a given video.}
\label{fig:teaser}
\vspace{-0.25in}
\end{figure}

Video corpus moment retrieval (VCMR)~\cite{escorcia2019finding,lei2020tvr} aims to find the video and time interval that match a text query in a corpus of videos, as in \cref{fig:teaser}.
VCMR involves two tasks: video retrieval (VR) and single-video moment retrieval (SVMR)~\cite{gao2017tall,anne2017localizing}.
VR retrieves candidate videos that may contain the described moment.
SVMR then predicts the time interval of the moment within a retrieved video.

VCMR is challenging because it requires both retrieving the correct video and localizing the correct moment within it~\cite{escorcia2019finding}.
For VR, a query describes only a small portion of a video, while much of its content is unrelated to the query~\cite{hou2024improving}.
Moreover, many videos in a corpus share similar activities.
Hence, recognizing the query alone is insufficient to identify the correct video.
Distinguishing these videos requires matching fine-grained details~\cite{chen2024verified}.
Even after retrieving the correct video, SVMR must identify the time interval of the matching moment.
This is challenging when similar queries occur close together, as frame-level embedding similarities may not distinguish their time intervals.
Moreover, benchmark annotations often include context before and after the moment that the query describes.
Hence, even a correctly located moment may not align with the annotated interval.

\begin{figure*}[t]
\centering
\includegraphics[width=\linewidth]{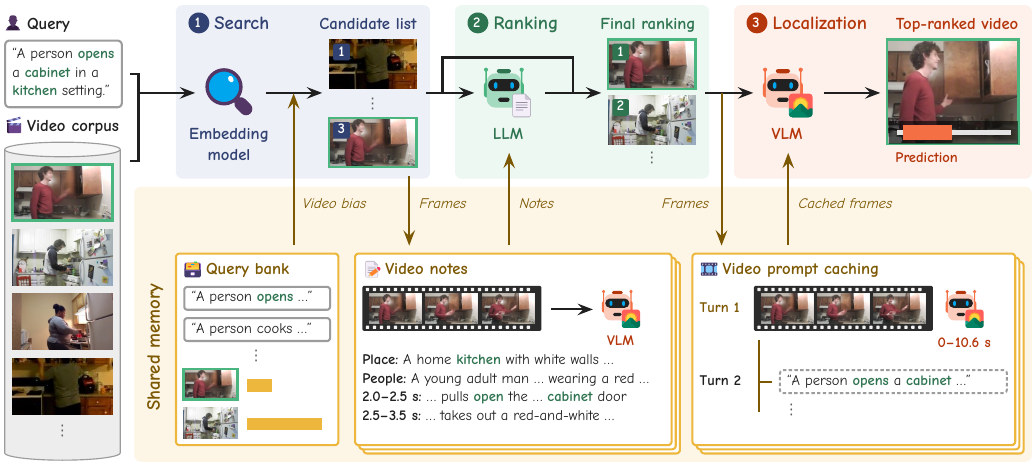}
\vspace{-0.22in}
\caption{\textbf{Pipeline of \method.}
The top row follows one query of a batch through the three stages.
The bottom row is the memory that the queries of the batch share.
\textcircled{\scriptsize 1} Search: the embedding model scores each video by its best-matching frame.
The query bank subtracts a bias from each score.
The bias is larger for a video that matches more queries.
\textcircled{\scriptsize 2} Ranking: a VLM writes notes on each candidate video once.
An LLM orders the candidates from their notes.
Reciprocal rank fusion merges this order with the embedding-based ranking.
\textcircled{\scriptsize 3} Localization: the VLM predicts the start and end of the moment in the top-ranked video.
Queries for the same video share a prompt prefix containing its timestamped frames. Each request adds the current query.
}
\label{fig:framework}
\vspace{-0.12in}
\end{figure*}

We present \method, a training-free framework for VCMR that processes a batch of queries over the same corpus together (\cref{fig:framework}).
\method has three stages: search, ranking, and localization.
Each stage uses a memory shared by all queries in the batch: a query bank, video notes, and video prompt cache, respectively.
In the search stage, an embedding model scores each video by its best-matching frame.
However, some videos may match many queries and outrank the correct one~\cite{radovanovic2010hubs,bogolin2022cross}.
To address this, \method penalizes each video by how strongly it matches the other queries in the batch~\cite{chowdhury2024nearest}.
In the ranking stage, \method reranks the candidate videos of each query.
However, frame-level similarity may miss the fine-grained details that separate similar videos~\cite{chen2024verified}.
To address this, we use a VLM to write notes on the scenes, people, objects, and actions of each candidate video.
Then, an LLM ranks the candidates from these notes.
The notes of a video are reused for every query whose candidates include it.
We use reciprocal rank fusion~\cite{cormack2009reciprocal} to combine this LLM ranking with the embedding-based ranking.
In the localization stage, we use a VLM to predict the time interval of the matching moment in the top-ranked video.
Multiple queries may select the same video.
We therefore use prompt caching to reduce the computational cost.

We evaluate \method on Charades-FIG, DiDeMo-FIG, and ActivityNet-FIG from VERIFIED~\cite{chen2024verified}.
Without training, \method retrieves the correct video more often than the trained VideoSearch-R1~\cite{lee2026videosearch} on all three benchmarks.
Without any labeled moment, \method performs similarly to VideoSearch-R1 in VCMR R@1 on ActivityNet-FIG, while underperforming at IoU 0.5 on Charades-FIG and DiDeMo-FIG.
The reason is that these benchmarks annotate broader intervals.
A margin of 2\,s per side widens the predicted interval to match these annotation-defined broader intervals (\cref{sec:vcmr_results}).
With this margin, \method outperforms VideoSearch-R1 in VCMR R@1 at IoU 0.5 on all three benchmarks.

Our contribution is threefold:
(I) Methodologically, we propose \method, a training-free framework for VCMR.
Its search, ranking, and localization stages uses a query bank, video notes, and a video prompt cache, respectively.
(II) Empirically, \method outperforms trained baselines in VR R@1 on the three VERIFIED benchmarks without training.
In VCMR, it performs similarly to the trained VideoSearch-R1 on ActivityNet-FIG.
With a single margin calibrated on a few labeled queries, it outperforms all trained baselines at IoU 0.5.
The query bank and the video notes each improve VR R@1 on all three test splits.
Reusing the notes and the cached frames of each video across queries reduces the input tokens of ranking and localization.
(III) Analytically, we show that the remaining gap in VCMR follows the annotation convention of the benchmarks.
Human annotators also mark shorter moments than the original ones, closer in length to those of \method. 
A trained model may learn this convention and score higher than the training-free method.

\section{Related Work}
\label{sec:related}

\noindent \textbf{Video Corpus Moment Retrieval.}
VCMR~\cite{escorcia2019finding,lei2020tvr} retrieves the video that matches a text query from a corpus and localizes the matching moment in it.
Early methods learn retrieval and localization from the annotations of each benchmark~\cite{lei2020tvr,li2020hero,zhang2021video,hou2021conquer,yoon2022selective,zhang2025video,hou2026event}, also in several languages~\cite{lei2021mtvr}.
VERIFIED~\cite{chen2024verified} rewrites the queries of Charades-STA~\cite{sigurdsson2016hollywood,gao2017tall}, DiDeMo~\cite{anne2017localizing}, and ActivityNet Captions~\cite{krishna2017dense} into fine-grained descriptions.
On Charades-FIG, five such models stay below 3\% VCMR R@1 at IoU 0.5.
Reinforcement learning has been used to navigate a ranked list with a user's feedback~\cite{ma2022interactive}.
VideoSearch-R1~\cite{lee2026videosearch} uses it to train a VLM that alternates between retrieval and localization while refining the query.
One training-free method captions the frames of each video and retrieves events by caption similarity~\cite{pan2025refcap}.
Another matches event chains that an LLM derives from the query against chains built from subtitles and keyframes~\cite{zhao2026training}.
For SVMR, trained models score candidate moments~\cite{zhang2020learning}, video LLMs predict timestamps as text~\cite{bai2025qwen3,ren2024timechat,wang2026time}, and training-free methods score sub-events or frames with frozen models~\cite{zheng2024training,huang2026your}.
The moment annotations of Charades-STA and ActivityNet Captions are biased in their temporal locations~\cite{otani2020uncovering,yuan2021closer}.
Annotators often disagree with them~\cite{otani2020uncovering}, and a re-annotation refines many~\cite{zhang2026timelens}.

\noindent \textbf{Video Retrieval and Reranking.}
VLM-based embedding models~\cite{jiang2024vlm2vec,li2026qwen3} embed videos and text queries in a shared space and rank the videos by their similarity to the query.
TF-PRVR~\cite{kim2026tf} retrieves partially relevant videos without training.
In such high-dimensional spaces, a few items, called hubs, are the nearest neighbors of many queries~\cite{radovanovic2010hubs,bogolin2022cross}.
QB-Norm~\cite{bogolin2022cross} renormalizes the similarities with a bank of training queries.
NNN~\cite{chowdhury2024nearest} subtracts from each candidate's score the mean similarity of its $k$ nearest queries in a reference database.
LLMs then rerank the retrieved passages listwise~\cite{sun2023chatgpt} or pairwise~\cite{qin2024large}.
The query bank of \method applies the correction of NNN, with the batch of input queries as the reference database.
It thus needs no training queries.
\method reranks its candidates listwise from their notes and fuses the result with the embedding-based ranking~\cite{cormack2009reciprocal}.

\noindent \textbf{Video Agents and Memory.}
Video agents answer a query about a long video~\cite{liu2025towards}.
They select keyframes~\cite{wang2024videoagent}, reason over captions~\cite{zhang2024simple} or trees of frame clusters~\cite{wang2025videotree}, query memories and tools~\cite{fan2024videoagent}, or retrieve
videos~\cite{jeong2025videorag}.
Some keep a compact summary across rounds~\cite{xu2026towards}.
However, those methods' memory is only for answering a query. 
For many queries regarding a single video, we need a memory shared across tasks, which appears in LLM agents and serving systems.
Agents keep reflections from earlier trials~\cite{shinn2023reflexion} or an evolving cheatsheet~\cite{suzgun2026dynamic}.
Serving systems return a stored response for a similar prompt~\cite{schroeder2025vcache} or reuse the cache of a shared prompt prefix~\cite{zheng2024sglang}.
\method shares memory across the queries of a batch.
It stores the queries, the notes of each video, and its frames.
Its video prompt caching applies prompt caching to the frames of each video.

\section{Methodology}
\label{sec:method}
\method processes a batch of text queries over a video corpus in three stages: search, ranking, and localization (\cref{fig:framework}).
All queries of the batch share one memory per stage: a query bank for search (\cref{sec:search}), video notes for ranking (\cref{sec:rank}), and a video prompt cache for localization (\cref{sec:localize}).

\noindent \textbf{Problem Setup.}
Let $\mathcal{V}=\{v_i\}_{i=1}^{N}$ denote a corpus of $N$ videos.
Each video $v$ has a duration of $T_v$ frames, and $x_v(t)$ denotes its frame at time $t$.
For each query $q$, the ground truth is a video $v^\star$ and a temporal interval $[t_s^\star,t_e^\star]$ within that video.
Here $t_s^\star$ and $t_e^\star$ denote the start and end times.
In VR, the videos are ranked by their relevance to the query $q$.
This produces a ranking $\pi=(\pi_1,\pi_2,\ldots,\pi_N)$, where $\pi_i$ denotes the video at rank $i$.
VR R@$k$ measures the fraction of queries for which the ground-truth video $v^\star$ appears among the top $k$ ranked videos.
In SVMR, the ground-truth video $v^\star$ is given.
R@1 at IoU threshold $\theta$ measures the fraction of queries whose predicted interval reaches a temporal IoU of at least $\theta$ with the ground truth.
VCMR combines both tasks and requires the correct video and the matching interval.
The predicted interval $[\hat{t}_s,\hat{t}_e]$ is within the top-ranked video $\pi_1$.
VCMR R@1 at IoU threshold $\theta$ measures the fraction of queries for which $\pi_1=v^\star$ and $\iou([\hat{t}_s,\hat{t}_e],[t_s^\star,t_e^\star])\ge\theta$.

\subsection{Search}
\label{sec:search}
We first use an image-text embedding model to retrieve candidate videos in $\mathcal{V}$ for each query $q$.
We sample each video at a fixed rate.
We define the video score $s(q,v)$ as the highest similarity between the query $q$ and its sampled frames:
\begin{equation}
s(q,v)=\max_{1\le t\le T_v}\cosine\big(q^{\text{emb}},x_v^{\text{emb}}(t)\big),
\label{eq:score}
\end{equation}
where $q^{\text{emb}}$ and $x_v^{\text{emb}}(t)$ denote the embeddings of the query $q$ and the frame $x_v(t)$, respectively.
However, videos that score highly for many queries can outrank the correct video $v^\star$~\cite{radovanovic2010hubs,bogolin2022cross}.
To reduce this bias, we adjust the video scores using a query bank $\mathcal{Q}$ that contains all input queries.

\noindent \textbf{Query Bank.}
For video $v$, we compute its similarity $s(q',v)$ with all queries $q'$ in the query bank $\mathcal{Q}$.
We take the average of its top $K_b$ similarity scores as the video-specific bias, as in nearest neighbor normalization~\cite{chowdhury2024nearest}:
\begin{equation}
b(v)=\frac{1}{K_b}\sum_{j=1}^{K_b}s_{(j)}(v),
\end{equation}
where $s_{(j)}(v)$ denotes the $j$-th highest similarity score for video $v$ across all queries in $\mathcal{Q}$.
We then adjust the video score as
\begin{equation}
\tilde{s}(q,v)=s(q,v)-\alpha b(v),
\label{eq:bank}
\end{equation}
where $\alpha$ controls the strength of the penalty.
Other queries, such as training queries~\cite{bogolin2022cross}, can also serve as $\mathcal{Q}$ (\cref{sec:ablations}).
Finally, we rank the videos by their adjusted scores $\tilde{s}(q,v)$ and select the top $K$ as the candidate list $\mathcal{C}(q)$.
We write $\mathcal{C}(q)=(c_1,\ldots,c_K)$, where $c_i$ is the video at rank $i$.

\subsection{Ranking} 
\label{sec:rank}
The search stage ranks videos by frame-level similarity, which may miss whether a video matches the query.
To refine the ranking, a VLM writes notes on each candidate video, and an LLM reranks the candidates from these notes.
We combine the result with the embedding-based ranking.

\noindent \textbf{Video Notes.}
For each candidate video $v\in\bigcup_{q\in\mathcal{Q}}\mathcal{C}(q)$, we use a VLM to write notes $\mathcal{N}(v)$ and reuse them for every query whose candidate list contains $v$.
The VLM receives uniformly sampled frames from video $v$:
\begin{equation}
\mathcal{N}(v)=\mathrm{VLM}\big(x_v(t_1),\ldots,x_v(t_{n_v})\big),
\label{eq:notes}
\end{equation}
where the $n_v$ frame indices $t_1<\cdots<t_{n_v}$ are evenly spaced in $[1,T_v]$.
The notes describe locations shown in the video and what is visible at each location.
They identify people, including their appearance and clothing, and describe notable objects by their colors and materials.
They also record actions in temporal order, specifying the start and end times of each action and the people and objects involved. 

\noindent \textbf{Notes-Based Ranking.} 
For each query $q$, we provide an LLM with the query and the notes of the $K$ candidate videos in $\mathcal{C}(q)$.
The LLM ranks these videos by their relevance to the query~\cite{sun2023chatgpt,qin2024large}:
\begin{equation}
(o_1,\ldots,o_K)=\mathrm{LLM}\big(q,\mathcal{N}(c_1),\ldots,\mathcal{N}(c_K)\big),
\label{eq:rank}
\end{equation}
where $o_i$ denotes the video at rank $i$ in the notes-based ranking.
The LLM compares each video's notes with the query, focusing on the described actions, people, and objects.
If the notes do not mention a detail in the query, the LLM does not assume it is absent from the video.
However, if the notes contradict the query, the LLM treats this as stronger evidence against the video. 
We combine the notes-based ranking with the embedding-based ranking using reciprocal rank fusion (RRF)~\cite{cormack2009reciprocal}.
For each candidate video $v\in\mathcal{C}(q)$, we denote its combined score by $r(v)$.
Let $\rho_{\text{notes}}(v)$ and $\rho_{\text{search}}(v)$ denote the notes-based rank and embedding-based rank of video $v$, so $\rho_{\text{notes}}(o_i)=\rho_{\text{search}}(c_i)=i$.
We compute the combined score as
\begin{equation}
r(v)=\frac{1}{k+\rho_{\text{notes}}(v)}+\frac{1}{k+\rho_{\text{search}}(v)},
\label{eq:fusion}
\end{equation}
where $k$ is the RRF constant.
Each ranking contributes a higher score to videos ranked closer to the top.
We then sort the candidate videos by $r(v)$.
The sorted candidates form the first $K$ videos $\pi_1,\ldots,\pi_K$ of the final ranking $\pi$.

\subsection{Localization}
\label{sec:localize}
For each query $q$, we use a VLM to predict the start and end times of the matching moment in the top-ranked video ${u=\pi_1}$.
We uniformly sample $m$ frames $x_u(t_1),\ldots,x_u(t_m)$ from the video, subject to the frame budget and sampling rate.
We provide these frames to the VLM with their timestamps.
The VLM then predicts the interval that matches $q$:
\begin{equation}
[\hat{t}_s,\hat{t}_e]=\mathrm{VLM}\big(q,x_u(t_1),\ldots,x_u(t_m)\big).
\label{eq:localize}
\end{equation}
For a long video, these frames are sparse.
We then refine the prediction with a second run on denser frames from $[\hat{t}_s-\Delta,\hat{t}_e+\Delta]$, where $\Delta=t_2-t_1$ is the frame spacing.

\noindent \textbf{Video Prompt Caching.}
Multiple queries may select the same video $u$ for localization.
To avoid repeatedly processing the same video frames, we use prompt caching.
For each video $u$, we send the sampled frames $x_u(t_1),\ldots,x_u(t_m)$ and their timestamps to the VLM and ask for the start and end of the video.
These frames, timestamps, and the initial response form a shared prompt prefix.
For each query $q$ assigned to $u$, we send this prefix followed by $q$ and ask the VLM to predict the matching interval $[\hat{t}_s,\hat{t}_e]$.
The requests use the same cache key for $u$, so the service can reuse the prefix across queries.
Each request contains only the current query and the shared video context.

\section{Experimental Studies}
\label{sec:exp}
\begin{table*}[t]
\centering
\caption{\textbf{VCMR on the VERIFIED Benchmarks} (test splits; search over the whole test corpus).
VR R@1 and R@10: the percentage of queries whose ground-truth video is among the top 1 and top 10 ranked videos.
VCMR R@1 at IoU 0.5 and 0.7: the percentage of queries whose top-ranked video is the ground-truth video and whose predicted interval has a temporal IoU of at least 0.5 or 0.7 with the ground-truth moment.
\method: the interval that the VLM predicts.
W/ margin: this interval widened by 2\,s on each side, a value picked with 20 labeled training queries.
Baselines as reported by VERIFIED~\cite{chen2024verified} and by \citet{lee2026videosearch}.
Qwen3-VL-2B: the zero-shot baseline of VideoSearch-R1, its VLM without fine-tuning.
Without training, \method achieves the highest VR R@1 on all three benchmarks.
It outperforms Qwen3-VL-2B, the other method without training, in every column, by about 10 points in VR R@1.
With the margin, it also achieves the highest VCMR R@1 at IoU 0.5 on all three benchmarks, and at 0.7 on all benchmarks except DiDeMo-FIG.}
\vspace{-0.08in}
\label{tab:vcmr}
\setlength{\tabcolsep}{3.4pt}
\resizebox{0.85\textwidth}{!}{%
\begin{tabular}{lcccccccccccccc}
\toprule
& & \multicolumn{4}{c}{Charades-FIG} & \multicolumn{4}{c}{DiDeMo-FIG} & \multicolumn{4}{c}{ActivityNet-FIG} \\
\cmidrule(lr){3-6}\cmidrule(lr){7-10}\cmidrule(lr){11-14}
& & \multicolumn{2}{c}{VR} & \multicolumn{2}{c}{VCMR R@1} & \multicolumn{2}{c}{VR} & \multicolumn{2}{c}{VCMR R@1} & \multicolumn{2}{c}{VR} & \multicolumn{2}{c}{VCMR R@1} \\
\cmidrule(lr){3-4}\cmidrule(lr){5-6}\cmidrule(lr){7-8}\cmidrule(lr){9-10}\cmidrule(lr){11-12}\cmidrule(lr){13-14}
Method & Trained & R@1 & R@10 & 0.5 & 0.7 & R@1 & R@10 & 0.5 & 0.7 & R@1 & R@10 & 0.5 & 0.7 \\
\midrule
HERO~\cite{li2020hero} & \checkmark & 1.69 & 11.51 & 0.11 & 0.05 & 8.48 & 39.52 & 0.24 & 0.17 & 7.95 & 36.49 & 1.46 & 0.75 \\
XML~\cite{lei2020tvr} & \checkmark & 2.80 & 14.11 & 1.05 & 0.43 & 14.83 & 53.95 & 3.19 & 2.32 & 13.46 & 49.99 & 2.81 & 1.63 \\
ReLoCLNet~\cite{zhang2021video} & \checkmark & 2.42 & 12.61 & 0.78 & 0.30 & 14.08 & 50.88 & 3.74 & 1.92 & 17.49 & 56.49 & 3.72 & 2.23 \\
CONQUER~\cite{hou2021conquer} & \checkmark & 2.80 & 14.11 & 1.21 & 0.65 & 14.83 & 53.95 & 5.48 & 3.66 & 13.46 & 49.99 & 2.95 & 1.63 \\
SQuiDNet~\cite{yoon2022selective} & \checkmark & 11.67 & 44.01 & 2.61 & 0.94 & 16.94 & 59.26 & 2.89 & 0.52 & 32.57 & 87.93 & 4.66 & 2.10 \\
VideoSearch-R1~\cite{lee2026videosearch} & \checkmark & 24.6 & 52.2 & 13.4 & 8.2 & 59.0 & 87.8 & 30.2 & \textbf{19.7} & 61.1 & \textbf{88.5} & 22.3 & 12.3 \\
\midrule
Qwen3-VL-2B~\cite{lee2026videosearch} & -- & 21.6 & 51.5 & 7.2 & 2.9 & 54.8 & 85.6 & 10.6 & 4.0 & 55.1 & 86.7 & 10.1 & 5.8 \\
\method & -- & \cellcolor{LightGray}\textbf{32.77} & \cellcolor{LightGray}\textbf{62.42} & \cellcolor{LightGray}11.75 & \cellcolor{LightGray}5.38 & \cellcolor{LightGray}\textbf{64.64} & \cellcolor{LightGray}\textbf{89.86} & \cellcolor{LightGray}14.88 & \cellcolor{LightGray}5.91 & \cellcolor{LightGray}\textbf{65.96} & \cellcolor{LightGray}87.60 & \cellcolor{LightGray}22.39 & \cellcolor{LightGray}13.31 \\
\method\ w/ margin & -- & \cellcolor{LightCyan}\textbf{32.77} & \cellcolor{LightCyan}\textbf{62.42} & \cellcolor{LightCyan}\textbf{21.37} & \cellcolor{LightCyan}\textbf{14.84} & \cellcolor{LightCyan}\textbf{64.64} & \cellcolor{LightCyan}\textbf{89.86} & \cellcolor{LightCyan}\textbf{32.44} & \cellcolor{LightCyan}16.37 & \cellcolor{LightCyan}\textbf{65.96} & \cellcolor{LightCyan}87.60 & \cellcolor{LightCyan}\textbf{24.91} & \cellcolor{LightCyan}\textbf{13.92} \\
\bottomrule
\end{tabular}}
\vspace{-0.12in}
\end{table*}

We evaluate \method on VCMR against trained and training-free baselines.
We also conduct ablation studies to analyze the accuracy and efficiency of our method.
Finally, we compare its predicted moments with human annotations.
\label{sec:vcmr_results}

\noindent\textbf{Benchmarks and Metrics.}
We evaluate on three benchmarks from VERIFIED~\cite{chen2024verified}:
Charades-FIG, DiDeMo-FIG, and ActivityNet-FIG.
They keep the videos, moments and splits of Charades-STA~\cite{gao2017tall}, DiDeMo~\cite{anne2017localizing} and ActivityNet Captions~\cite{krishna2017dense}.
VERIFIED replaces each query by a fine-grained description of its moment, generated by LLMs and multimodal models and filtered for inaccurate details.
These queries are more than twice as long as the original ones.
Charades-FIG has 3,720 test queries over 1,334 Charades videos of indoor daily activities~\cite{sigurdsson2016hollywood}.
DiDeMo-FIG has 4,013 test queries over 1,002 personal videos from Flickr.
ActivityNet Captions has no public test annotations and annotates its validation videos twice.
VERIFIED builds the ActivityNet-FIG test split from the second annotation, with 16,966 queries over 4,866 YouTube videos of human activities.
These videos average 118\,s and their moments 40\,s, against 30\,s and 8\,s on Charades-FIG.
Each video has about three to four test queries.
We search the entire video corpus of each test split and process all its queries as a batch.
We report VR R@1 and R@10, and VCMR R@1 at temporal IoU thresholds $\theta$ of 0.5 and 0.7, as defined in \cref{sec:method}.

\noindent \textbf{Implementation Details.}
We use Qwen3-VL-Embedding-2B~\cite{li2026qwen3} for search and gpt-6-luna for both the VLM and LLM.
In the search stage, we embed the queries and the frames of each video, sampled at 2 fps.
We select the top $K=10$ videos.
The query bank $\mathcal{Q}$ contains all queries in the evaluated split, with $K_b=20$ and $\alpha=0.75$.
In the ranking stage, the VLM writes the notes of a video from at most 64 frames at up to 2 fps.
We then combine the notes-based and embedding-based rankings by RRF with $k=60$~\cite{cormack2009reciprocal}.
In the localization stage, we sample at most 120 frames from the top-ranked video $\pi_1$ at up to 4 fps.

\noindent \textbf{Margin.}
We report every result both without and with a fixed margin around the predicted interval.
Benchmark annotations often add context before and after the action that the query describes.
Trained models learn this convention from the training split.
\method does not learn this convention as it is not trained on any annotation.
Hence, its interval can locate the action correctly yet be shorter than the annotation.
Human annotators show the same gap (\cref{sec:analysis}).
The margin extends $[\hat{t}_s,\hat{t}_e]$ by $w$ seconds on each side to $[\hat{t}_s-w,\hat{t}_e+w]$.
We set $w=2$s for all three benchmarks.
We pick this margin using only 20 labeled queries from the training set. .

\noindent \textbf{Baselines.}
We compare with five trained VCMR models on the VERIFIED benchmark.
HERO~\cite{li2020hero} encodes each frame with a cross-modal Transformer and the whole video with a temporal Transformer.
XML~\cite{lei2020tvr} encodes the query and the video separately and detects the start and end of the moment with 1D convolution filters on their similarity curve.
ReLoCLNet~\cite{zhang2021video} also encodes them separately and aligns them by contrastive learning at the video and frame levels.
CONQUER~\cite{hou2021conquer} couples the query and the video by bi-directional attention and localizes the moment within the top videos of XML.
SQuiDNet~\cite{yoon2022selective} reranks these videos and debiases its predictions selectively, guided by the query's meaning.
VideoSearch-R1~\cite{lee2026videosearch} trains Qwen3-VL-2B with supervised fine-tuning and reinforcement learning to verify the retrieved video, refine the query, and localize the moment.
VideoSearch-R1 searches with Qwen3-VL-Embedding-2B, as our method.
When the retrieved video does not match the query, it refines the query and searches again.
Its zero-shot baseline uses Qwen3-VL-2B without fine-tuning and never refines the query.
When a video does not match, it moves on to the next video in the same ranking.
It localizes the moment in the first video that it judges to match the query.
\Cref{tab:vcmr} reports their published numbers.
In \method, the embedding model only retrieves videos.
Our embedding-only baseline also uses it to predict the moment.
It ranks the videos by their best-matching frame, without the query bank.
In its top-ranked video, it predicts the half-height moment.
This moment spans the frames around the best-matching frame that score above the midpoint of the video's scores.
\begin{figure*}[t]
\centering
\includegraphics[width=0.99\linewidth]{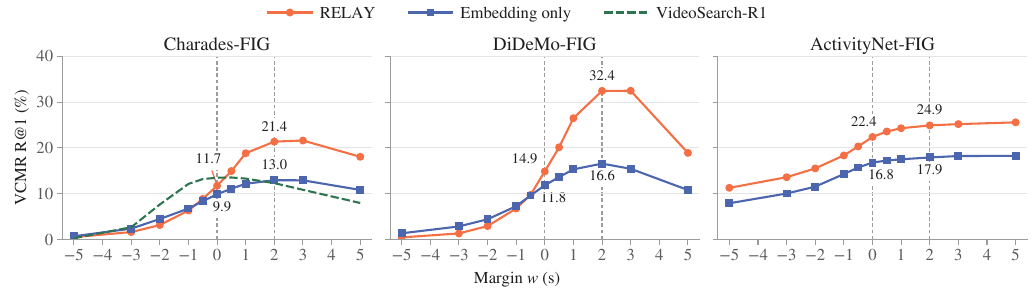}
\vspace{-0.22in}
\caption{\textbf{Margin around the predicted interval}: VCMR R@1 (\%) at IoU 0.5 on the three test splits when the predicted interval is widened by $w$ seconds on each side, or shrunk for $w<0$.
Gray dashed lines: $w=0$ and $w=2$\,s.
Embedding only: our baseline that uses only Qwen3-VL-Embedding-2B, the embedding model of \method\ and VideoSearch-R1~\cite{lee2026videosearch}, to retrieve the video and predict the moment.
Its moment grows from the best-matching frame of its top-ranked video while the frames score above the midpoint of the video's scores.
VideoSearch-R1: its released predictions on Charades-FIG, widened the same way.
Numbers: R@1 at $w=0$ and $w=2$\,s.
\method\ outperforms the embedding-only baseline at every $w\ge0$ on all three benchmarks.
The 2\,s margin raises \method\ and lowers VideoSearch-R1.}
\label{fig:test_margin}
\vspace{-0.12in}
\end{figure*}

\subsection{Main Results}
\label{sec:main_results}
Without training, \method achieves the highest VR R@1 on all three benchmarks (\cref{tab:vcmr}).
On Charades-FIG, it reaches 32.8\%, against 24.6\% for the trained VideoSearch-R1.
Video notes contribute the most to this lead (\cref{sec:ablations}).
Without any labeled moment, \method outperforms the zero-shot Qwen3-VL-2B in VCMR R@1 on all three benchmarks (\cref{tab:vcmr}).
It performs similarly to VideoSearch-R1 on ActivityNet-FIG (22.4\% vs.\ 22.3\% at IoU 0.5) but falls behind it on Charades-FIG and DiDeMo-FIG.
This gap follows how the benchmarks annotate a moment.
\method's predicted interval lies where human annotators place the moment, but it is shorter than the annotation (\cref{sec:analysis}).
Charades-STA adds seconds before and after the action that the query names, and DiDeMo annotates moments as runs of 5-second segments.
VideoSearch-R1 is trained on these annotations and predicts longer moments.
On ActivityNet-FIG, whose moments are long, a few seconds change the IoU little.
With the 2s margin, \method beats all baselines in VCMR R@1 at IoU 0.5 on all three benchmarks and at 0.7 on all but DiDeMo-FIG.

\subsection{Ablation Studies}
\label{sec:ablations}
\begin{table}[t]
\centering
\caption{\textbf{Ablations}.
W/ margin set to the 2\,s margin.
W/o video prompt caching: one localization call per query.
Each of the three rows below it also uses one call per query and removes one more component.
W/ frames for ranking: one VLM call orders the ten candidates on their frames in place of the notes-based ranking.
Tokens: input tokens per query not read from the prompt cache, of ranking and of localization; frames: frames shown to the VLM per query; --: not counted.
The embedding-only baseline gives 25.97, 12.96 and 7.31, its half-height moment widened the same way.}
\label{tab:test_ablation}
\vspace{-0.08in}
\setlength{\tabcolsep}{2.4pt}
\resizebox{\columnwidth}{!}{%
\begin{tabular}{lcccccc}
\toprule
& VR & \multicolumn{2}{c}{VCMR R@1} & \multicolumn{2}{c}{Tokens} & \multirow{2}{*}{Frames} \\
\cmidrule(lr){3-4}\cmidrule(lr){5-6}
& R@1 & 0.5 & 0.7 & ranking & localization & \\
\midrule
\method\ w/ margin & 32.77 & 21.37 & 14.84 & 10.5k & 7.8k & 54.8 \\
\hphantom{\method\ }w/o margin & 32.77 & 11.75 & 5.38 & 10.5k & 7.8k & 54.8 \\
\hphantom{\method\ }w/o video prompt caching & 32.77 & 20.51 & 14.11 & 10.5k & 22.8k & 129.3 \\
\hphantom{\method\ }w/o fusion & 32.04 & 20.22 & 13.44 & 10.5k & 22.7k & 129.0 \\
\hphantom{\method\ }w/o query bank & 29.68 & 18.74 & 12.69 & 10.4k & 22.5k & 128.0 \\
\hphantom{\method\ }w/o notes & 28.36 & 17.58 & 12.20 & 0.0k & 22.7k & 108.5 \\
\hphantom{\method\ }w/ frames for ranking & 26.53 & 15.91 & 10.43 & 24.2k & 22.7k & 228.4 \\
\bottomrule
\end{tabular}}
\vspace{-0.1in}
\end{table}

\noindent \textbf{Effect of the Margin.}
The margin helps \method mainly because many of its predicted intervals are too short to reach IoU 0.5 against the annotation (\cref{fig:test_margin}).
A margin of $w=2$s raises VCMR R@1 at IoU 0.5 on all three test splits.
The gain is largest on DiDeMo-FIG (17.6 points) and smallest on ActivityNet-FIG, whose annotated moments are long.
We focus on the queries for which \method ranks the annotated video first.
For around 2/5 of these queries, the predicted interval is less than half as long as the annotated moment.
Such an interval stays below IoU 0.5 even when it lies inside the moment.
Most queries that reach IoU 0.5 only with the margin have such a short interval.
The margin only widens the interval in the top-ranked video.
Hence, it helps only when the retrieved video is correct.  
On Charades-FIG, the margin does not help VideoSearch-R1, which is trained on the benchmark.
However, the margin raises the embedding-only baseline, which is training-free like \method, on all three benchmarks.
Hence, we conclude that the training-free methods are underestimated due to the original annotation. 

\noindent \textbf{Component Analysis.} 
We analyze each component's effects, as their efficiency and accuracy in \cref{tab:test_ablation}. 
Video notes contribute the most to accuracy, and video prompt caching the most to efficiency.
With localization per query, removing the notes lowers VR R@1 from 32.8\% to 28.4\%.
The query bank and rank fusion add smaller gains in both VR and VCMR. 
With only the embedding model, VCMR R@1 at IoU 0.5 with the margin drops by 7.0 to 15.9 points on the three benchmarks (\cref{fig:test_margin}).
Using a VLM to direct ordering based on frames instead of notes increases the cost and decreases the accuracy.
Its VR R@1 is even below the 28.4\% without the notes.
It also reads twice more input tokens than ranking with the notes.
Since the notes are written once per video and shared by all its queries, their cost decreases as more queries come up with the same video.
Video prompt caching reduces the localization input tokens from 22.8k to 7.8k per query and also slightly raises VCMR R@1.

\begin{table}[t]
\centering
\caption{\textbf{Source of the Query Bank} (\%; test splits).
Test queries: $\mathcal{Q}$ holds all test queries of the split, as in \cref{tab:vcmr}.
Training queries: $\mathcal{Q}$ holds the queries of the benchmark's training split instead, with the rest of \method\ unchanged.
W/ margin: the predicted interval widened by 2\,s on each side, as in \cref{tab:vcmr}.
With training queries, every number changes by at most 2.2 points.}
\label{tab:test_bank}
\vspace{-0.08in}
\setlength{\tabcolsep}{3pt}
\resizebox{\columnwidth}{!}{%
\begin{tabular}{lccccccccc}
\toprule
& \multicolumn{3}{c}{Charades-FIG} & \multicolumn{3}{c}{DiDeMo-FIG} & \multicolumn{3}{c}{ActivityNet-FIG} \\
\cmidrule(lr){2-4}\cmidrule(lr){5-7}\cmidrule(lr){8-10}
& VR & \multicolumn{2}{c}{VCMR R@1} & VR & \multicolumn{2}{c}{VCMR R@1} & VR & \multicolumn{2}{c}{VCMR R@1} \\
\cmidrule(lr){3-4}\cmidrule(lr){6-7}\cmidrule(lr){9-10}
Query bank & R@1 & 0.5 & 0.7 & R@1 & 0.5 & 0.7 & R@1 & 0.5 & 0.7 \\
\midrule
Test queries & 32.77 & 11.75 & 5.38 & 64.64 & 14.88 & 5.91 & 65.96 & 22.39 & 13.31 \\
\quad w/ margin & 32.77 & 21.37 & 14.84 & 64.64 & 32.44 & 16.37 & 65.96 & 24.91 & 13.92 \\
\midrule
Training queries & 32.93 & 11.67 & 5.19 & 62.47 & 14.48 & 5.83 & 64.88 & 22.28 & 13.38 \\
\quad w/ margin & 32.93 & 21.26 & 14.95 & 62.47 & 31.27 & 15.50 & 64.88 & 24.78 & 13.82 \\
\bottomrule
\end{tabular}}
\vspace{-0.1in}
\end{table}

\begin{figure*}[t]
\centering
\includegraphics[width=0.99\linewidth]{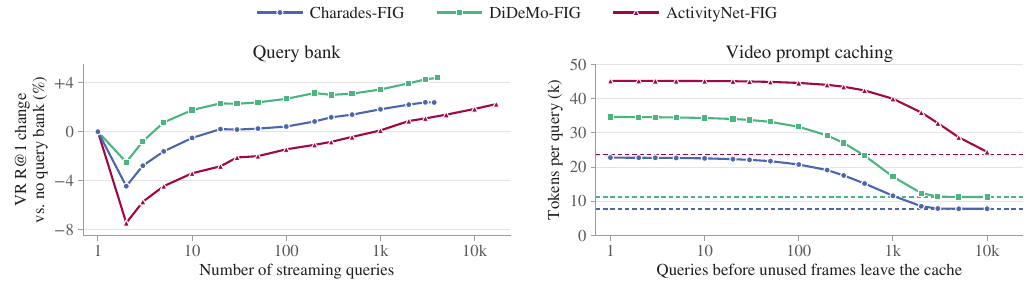}
\vspace{-0.22in}
\caption{\textbf{Streaming Queries}.
Query bank: VR R@1 of the search minus that without a query bank, for a query whose bank holds the given number of streaming queries, itself included; with all of them, the bank is the batch.
Video prompt caching: localization input tokens per query, by the number of queries for which the cache keeps unused frames (mean of 5 random orders).
Dashed lines: all queries in one batch.
A bank of a few queries lowers VR R@1, and ActivityNet-FIG needs the most queries for both.}
\label{fig:stream}
\vspace{-0.12in}
\end{figure*}

\noindent \textbf{Source of Query Bank Analysis.}
We test whether the queries in the query bank can come from the training set instead of the test set.
By default, $\mathcal{Q}$ holds all test queries of a split.
As in QB-Norm~\cite{bogolin2022cross}, we instead fill $\mathcal{Q}$ with the queries of each benchmark's training split.
\method then changes by at most 0.2 points on Charades-FIG.
Its largest loss is 2.2 points of VR R@1, on DiDeMo-FIG (\cref{tab:test_bank}).
Hence, whether the query bank comes from training and testing does not matter.
Videos that score highly for test queries also score highly for training queries.

\begin{table}[t]
\centering
\caption{\textbf{Streaming Queries} (\%; test splits).
Batch: all test queries of the split arrive together, as in \cref{tab:vcmr}.
Stream: the queries arrive one at a time in a random order, and $\mathcal{Q}$ holds those that have arrived.
W/ margin: the predicted interval widened by 2\,s on each side, as in \cref{tab:vcmr}.
In the stream, every number changes by at most 0.9 points.}
\label{tab:test_stream}
\vspace{-0.08in}
\setlength{\tabcolsep}{3pt}
\resizebox{\columnwidth}{!}{%
\begin{tabular}{lccccccccc}
\toprule
& \multicolumn{3}{c}{Charades-FIG} & \multicolumn{3}{c}{DiDeMo-FIG} & \multicolumn{3}{c}{ActivityNet-FIG} \\
\cmidrule(lr){2-4}\cmidrule(lr){5-7}\cmidrule(lr){8-10}
& VR & \multicolumn{2}{c}{VCMR R@1} & VR & \multicolumn{2}{c}{VCMR R@1} & VR & \multicolumn{2}{c}{VCMR R@1} \\
\cmidrule(lr){3-4}\cmidrule(lr){6-7}\cmidrule(lr){9-10}
Queries & R@1 & 0.5 & 0.7 & R@1 & 0.5 & 0.7 & R@1 & 0.5 & 0.7 \\
\midrule
Batch & 32.77 & 11.75 & 5.38 & 64.64 & 14.88 & 5.91 & 65.96 & 22.39 & 13.31 \\
\quad w/ margin & 32.77 & 21.37 & 14.84 & 64.64 & 32.44 & 16.37 & 65.96 & 24.91 & 13.92 \\
\midrule
Stream & 32.15 & 11.80 & 5.35 & 63.72 & 14.88 & 5.66 & 65.09 & 22.27 & 13.27 \\
\quad w/ margin & 32.15 & 21.24 & 14.84 & 63.72 & 32.07 & 16.00 & 65.09 & 24.67 & 13.77 \\
\bottomrule
\end{tabular}}
\vspace{-0.1in}
\end{table}

\subsection{Streaming Queries}
\label{sec:stream}
\begin{figure*}[t]
\centering
\includegraphics[width=\linewidth]{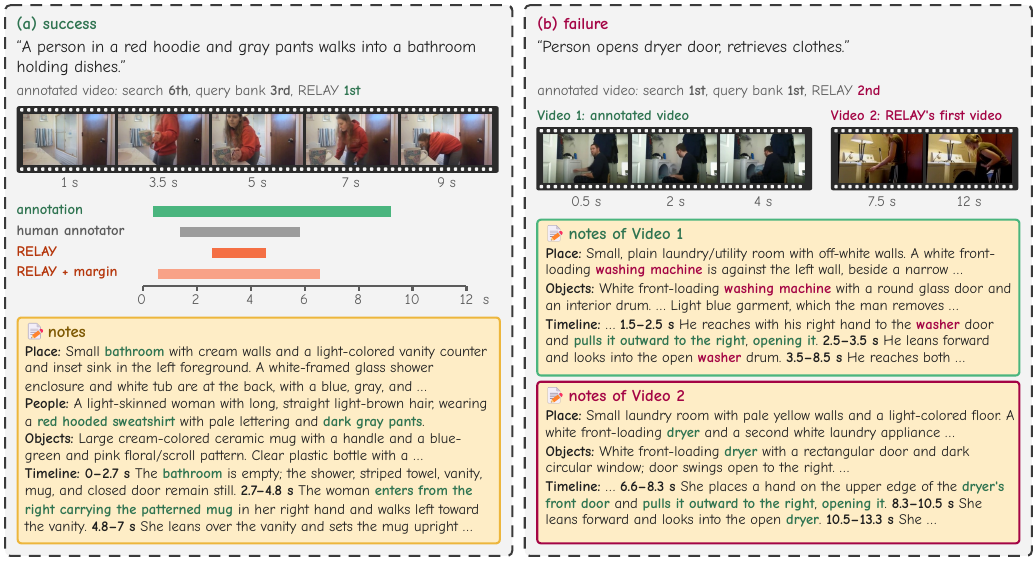}
\vspace{-0.22in}
\caption{\textbf{Qualitative Example.}
(a) Success.
The search ranks the annotated video 6th, the query bank raises it to 3rd, and \method ranks it first.
Its notes name the bathroom, the red hooded sweatshirt and the dark gray pants of the query.
They also record that the woman enters carrying a mug at 2.7--4.8\,s.
The predicted interval, 2.6--4.6\,s, lies within the moment of the human annotator, 1.4--5.8\,s.
The annotation, 0.4--9.2\,s, adds seconds on both sides.
The margin widens the predicted interval to 0.6--6.6\,s, closer to the annotation.
(b) Failure.
The search and the query bank rank the annotated Video~1 first.
Its notes describe a man opening the door and taking out clothes, but they call the machine a washing machine.
The notes of Video~2 describe a woman opening the door of a dryer, so \method ranks Video~2 first.
Bars: dataset annotation, human annotation, the predicted interval, and the interval with the 2\,s margin.
Human annotation is from \citet{otani2020uncovering}.}
\label{fig:examples}
\vspace{-0.12in}
\end{figure*}

We also explore the settings where queries may come in a streaming manner. 
In the streaming manner, queries arrive one at a time and accumulate in a random order.
Each query is answered before the next one arrives.
We construct the query bank $\mathcal{Q}$ only based on the queries that have arrived.
The video notes keep the same setting. 
The ranking stage then reranks each new candidate list $\mathcal{C}(q)$.
We simulate a cache that retains a video's frames for a fixed number of query arrivals after their last use.
A query reads its video's frames from the cache if the last query with the same top-ranked video came at most that many queries before it.
Otherwise, the VLM processes the video's frames again from scratch.

\noindent \textbf{Comparison.}
We compare streaming settings to our original settings. 
For the streaming setting, we run \method on one random order of each test split.
VR R@1 drops by 0.6 to 0.9 points (\cref{tab:test_stream}).
VCMR R@1 at IoU 0.5 changes by at most 0.4 points, with and without the margin.
Hence, \method can fit both streaming and non-streaming settings.

\noindent \textbf{Query Bank.}
In a stream, a query sees a smaller query bank than in the batch, and a small bank can hurt the search (\cref{fig:stream}).
With a few queries in the bank, the bias $b(v)$ depends on which queries happened to arrive, not on how many queries match video $v$.
VR R@1 of this setting then falls behind that without a bank.
As more queries arrive, $b(v)$ approaches its value in the batch.
The bank starts to help after 5 to 20 queries on Charades-FIG and DiDeMo-FIG, but only after about 1,000 on ActivityNet-FIG.
As the number of queries increases, VR R@1 tends to be stable and achieves similar performance to batch settings (\cref{tab:test_stream}).

\noindent \textbf{Video Prompt Caching.}
In a stream, caching saves tokens only if a video comes back before its frames leave the cache (\cref{fig:stream}).
In the batch, we group queries by video so successive requests can reuse the cached frames.
With a cache that keeps unused frames for 1,000 queries, localization on Charades-FIG reads 11.6k input tokens per query, against 7.8k in the batch and 22.8k without caching.
The stream saves 1/3 as many tokens as the batch on Charades-FIG and DiDeMo-FIG, but 1/4 on ActivityNet-FIG.
Its cache must keep frames five times longer for the same share.

\subsection{Analysis}
\label{sec:analysis}
\begin{table}[t]
\centering
\caption{\textbf{Benchmark and Human  Annotation}. 
We test R@1 (\%) at IoU 0.5 and 0.7. 
We test human annotation (one annotator), benchmark annotation, and \method against references. 
The references are the benchmark's annotation and the median moment of four of the five annotators of \citet{otani2020uncovering}. 
Offset: median distance (s) of the centers; length: median (s).}
\label{tab:label_humans}
\vspace{-0.08in}
\footnotesize
\setlength{\tabcolsep}{4pt}
\begin{tabular}{lcccccc}
\toprule
& \multicolumn{2}{c}{vs.\ annotation} & \multicolumn{2}{c}{vs.\ annotators} & & \\
\cmidrule(lr){2-3}\cmidrule(lr){4-5}
Moment & 0.5 & 0.7 & 0.5 & 0.7 & offset & length \\
\midrule
\multicolumn{7}{l}{Charades-STA} \\
Benchmark Annotation & -- & -- & 49.7 & 17.9 & 1.0 & 7.1 \\
Human annotator & 44.0 & 18.0 & 67.2 & 37.6 & 0.7 & 4.2 \\
\quad + 2\,s margin & 73.7 & 51.6 & 51.6 & 18.3 & 0.8 & 7.8 \\
\method & 23.4 & 11.6 & 43.2 & 22.9 & 0.8 & 2.9 \\
\method\ w/ margin & 65.6 & 46.2 & 57.2 & 21.3 & 0.9 & 6.5 \\
\midrule
\multicolumn{7}{l}{ActivityNet Captions} \\
Benchmark Annotation & -- & -- & 42.9 & 24.7 & 7.6 & 28.2 \\
Human annotator & 35.1 & 20.9 & 52.0 & 34.1 & 5.3 & 16.7 \\
\quad + 2\,s margin & 37.6 & 21.6 & 52.9 & 31.0 & 5.2 & 19.9 \\
\method & 30.0 & 17.9 & 43.9 & 29.7 & 5.0 & 9.6 \\
\method\ w/ margin & 33.6 & 20.3 & 51.4 & 32.8 & 4.9 & 13.2 \\
\bottomrule
\end{tabular}
\vspace{-0.2in}
\end{table}

\noindent \textbf{Human Annotations.}
\method's predicted interval is consistent with human annotation but is shorter than the benchmark annotation (\cref{tab:label_humans}).
\citet{otani2020uncovering} uses five annotators to re-mark the moments of 1,000 Charades-STA and 1,288 ActivityNet Captions test queries.
We score each moment against the annotation and the median moment of four annotators, so an annotator is never scored against itself.
On Charades-STA, an annotator agrees with the other annotators more than with the benchmark annotation (67.2\% vs.\ 44.0\% R@1 at IoU 0.5).
The benchmark annotation is longer than human annotation (7.1\,s vs.\ 4.2\,s).
Charades-STA reuses the interval of the matching Charades action label, which adds seconds around the action.
Setting human annotation as the baseline, \method beats the benchmark annotation at IoU 0.7 (22.9\% vs.\ 17.9\%).
The 2\,s margin lengthens \method's interval from 2.9\,s to 6.5\,s, which more fits to the dataset settings.
On an annotator's moment, the same margin raises R@1 against the annotation to 73.7\% but lowers it against the other annotators to 51.6\%.
Hence, the margin only fits the annotation convention.
On ActivityNet Captions, the 2\,s margin raises \method's R@1 at IoU 0.5 against the annotation only from 30.0\% to 33.6\%.

\noindent \textbf{Qualitative Results.}
We show the qualitative results, including both success and failure cases,  in \cref{fig:examples}.
In the success case, the query bank raises the annotated video from 6th to 3rd, and the notes name the red sweatshirt and the gray pants of the query.
In the failure case, they describe a washing machine, while the query and the notes of another video name a dryer, and \method ranks that other video first.

\vspace{-0.1in}
\section{Conclusion}
\vspace{-0.1in}

\label{sec:conclusion}
We present \method, a training-free framework for video corpus moment retrieval.
It has three stages: search, ranking, and localization, with each stage having its own memory mechanism.
In the search stage, it uses a query bank to penalize videos that score highly for many queries.
In the ranking stage, it uses video notes, written once per video, to rerank the candidates.
In the localization stage, it uses video prompt caching to reuse the frames of a video across its queries.
Without training, \method achieves a higher VR R@1 than the baselines.
Its predicted intervals are shorter than the benchmark annotations, but align with human annotations.
A single margin calibrated on a few labeled queries closes the gap between \method and benchmark annotations.

{
    \small
    \bibliographystyle{ieeenat_fullname}
    \bibliography{main}

\begin{thebibliography}{55}
\providecommand{\natexlab}[1]{#1}
\providecommand{\url}[1]{\texttt{#1}}
\expandafter\ifx\csname urlstyle\endcsname\relax
  \providecommand{\doi}[1]{doi: #1}\else
  \providecommand{\doi}{doi: \begingroup \urlstyle{rm}\Url}\fi

\bibitem[Anne~Hendricks et~al.(2017)Anne~Hendricks, Wang, Shechtman, Sivic, Darrell, and Russell]{anne2017localizing}
Lisa Anne~Hendricks, Oliver Wang, Eli Shechtman, Josef Sivic, Trevor Darrell, and Bryan Russell.
\newblock Localizing moments in video with natural language.
\newblock In \emph{Proceedings of the IEEE international conference on computer vision}, pages 5803--5812, 2017.

\bibitem[Bai et~al.(2025)Bai, Cai, Chen, Chen, Chen, Cheng, Deng, Ding, Gao, Ge, et~al.]{bai2025qwen3}
Shuai Bai, Yuxuan Cai, Ruizhe Chen, Keqin Chen, Xionghui Chen, Zesen Cheng, Lianghao Deng, Wei Ding, Chang Gao, Chunjiang Ge, et~al.
\newblock Qwen3-vl technical report.
\newblock \emph{arXiv preprint arXiv:2511.21631}, 2025.

\bibitem[Bogolin et~al.(2022)Bogolin, Croitoru, Jin, Liu, and Albanie]{bogolin2022cross}
Simion-Vlad Bogolin, Ioana Croitoru, Hailin Jin, Yang Liu, and Samuel Albanie.
\newblock Cross modal retrieval with querybank normalisation.
\newblock In \emph{2022 IEEE/CVF Conference on Computer Vision and Pattern Recognition (CVPR)}, pages 5184--5195. IEEE, 2022.

\bibitem[Chen et~al.(2024)Chen, Wang, Chen, Zhang, Feng, Huang, Jia, and Zhu]{chen2024verified}
Houlun Chen, Xin Wang, Hong Chen, Zeyang Zhang, Wei Feng, Bin Huang, Jia Jia, and Wenwu Zhu.
\newblock Verified: A video corpus moment retrieval benchmark for fine-grained video understanding.
\newblock \emph{Advances in Neural Information Processing Systems}, 37:\penalty0 40393--40406, 2024.

\bibitem[Chowdhury et~al.(2024)Chowdhury, Wang, Shenoy, Kiela, Schwettmann, and Thrush]{chowdhury2024nearest}
Neil Chowdhury, Franklin Wang, Sumedh Shenoy, Douwe Kiela, Sarah Schwettmann, and Tristan Thrush.
\newblock Nearest neighbor normalization improves multimodal retrieval.
\newblock In \emph{Proceedings of the 2024 Conference on Empirical Methods in Natural Language Processing}, pages 22571--22582, 2024.

\bibitem[Cormack et~al.(2009)Cormack, Clarke, and Buettcher]{cormack2009reciprocal}
Gordon~V Cormack, Charles~LA Clarke, and Stefan Buettcher.
\newblock Reciprocal rank fusion outperforms condorcet and individual rank learning methods.
\newblock In \emph{Proceedings of the 32nd international ACM SIGIR conference on Research and development in information retrieval}, pages 758--759, 2009.

\bibitem[Escorcia et~al.(2019)Escorcia, Soldan, Sivic, Ghanem, and Russell]{escorcia2019finding}
Victor Escorcia, Mattia Soldan, Josef Sivic, Bernard Ghanem, and Bryan Russell.
\newblock Finding moments in video collections using natural language.
\newblock \emph{arXiv preprint arXiv:1907.12763}, 2019.

\bibitem[Fan et~al.(2024)Fan, Ma, Wu, Du, Li, Gao, and Li]{fan2024videoagent}
Yue Fan, Xiaojian Ma, Rujie Wu, Yuntao Du, Jiaqi Li, Zhi Gao, and Qing Li.
\newblock Videoagent: A memory-augmented multimodal agent for video understanding.
\newblock In \emph{European Conference on Computer Vision}, pages 75--92. Springer, 2024.

\bibitem[Gao et~al.(2017)Gao, Sun, Yang, and Nevatia]{gao2017tall}
Jiyang Gao, Chen Sun, Zhenheng Yang, and Ram Nevatia.
\newblock Tall: Temporal activity localization via language query.
\newblock In \emph{2017 IEEE International Conference on Computer Vision (ICCV)}, pages 5277--5285. IEEE, 2017.

\bibitem[Hou et~al.(2023)Hou, Pang, Lan, Shen, and Cheng]{hou2023multi}
Danyang Hou, Liang Pang, Yanyan Lan, Huawei Shen, and Xueqi Cheng.
\newblock Multi-video moment ranking with multimodal clue.
\newblock \emph{arXiv preprint arXiv:2301.13606}, 2023.

\bibitem[Hou et~al.(2024)Hou, Pang, Shen, and Cheng]{hou2024improving}
Danyang Hou, Liang Pang, Huawei Shen, and Xueqi Cheng.
\newblock Improving video corpus moment retrieval with partial relevance enhancement.
\newblock In \emph{Proceedings of the 2024 International Conference on Multimedia Retrieval}, pages 394--403, 2024.

\bibitem[Hou et~al.(2026)Hou, Pang, Lan, Shen, and Cheng]{hou2026event}
Danyang Hou, Liang Pang, Yanyan Lan, Huawei Shen, and Xueqi Cheng.
\newblock Event-aware video corpus moment retrieval.
\newblock In \emph{European Conference on Information Retrieval}, pages 3--17. Springer, 2026.

\bibitem[Hou et~al.(2021)Hou, Ngo, and Chan]{hou2021conquer}
Zhijian Hou, Chong-Wah Ngo, and Wing~Kwong Chan.
\newblock Conquer: Contextual query-aware ranking for video corpus moment retrieval.
\newblock In \emph{Proceedings of the 29th ACM International Conference on Multimedia}, pages 3900--3908, 2021.

\bibitem[Huang et~al.(2026)Huang, Devereux, and Wang]{huang2026your}
Ji Huang, Barry Devereux, and Hui Wang.
\newblock Your vlm already knows when: Training-free temporal grounding by asking yes or no.
\newblock \emph{arXiv preprint arXiv:2608.08315}, 2026.

\bibitem[Jeong et~al.(2025)Jeong, Kim, Baek, and Hwang]{jeong2025videorag}
Soyeong Jeong, Kangsan Kim, Jinheon Baek, and Sung~Ju Hwang.
\newblock Videorag: Retrieval-augmented generation over video corpus.
\newblock In \emph{Findings of the Association for Computational Linguistics: ACL 2025}, pages 21278--21298, 2025.

\bibitem[Jiang et~al.(2024)Jiang, Meng, Yang, Yavuz, Zhou, and Chen]{jiang2024vlm2vec}
Ziyan Jiang, Rui Meng, Xinyi Yang, Semih Yavuz, Yingbo Zhou, and Wenhu Chen.
\newblock Vlm2vec: Training vision-language models for massive multimodal embedding tasks.
\newblock \emph{arXiv preprint arXiv:2410.05160}, 2024.

\bibitem[Kim and Eom(2026)]{kim2026tf}
Giyeol Kim and Chanho Eom.
\newblock Tf-prvr: Training-free partially relevant video retrieval.
\newblock \emph{arXiv preprint arXiv:2610.07925}, 2026.

\bibitem[Krishna et~al.(2017)Krishna, Hata, Ren, Fei-Fei, and Carlos~Niebles]{krishna2017dense}
Ranjay Krishna, Kenji Hata, Frederic Ren, Li Fei-Fei, and Juan Carlos~Niebles.
\newblock Dense-captioning events in videos.
\newblock In \emph{Proceedings of the IEEE international conference on computer vision}, pages 706--715, 2017.

\bibitem[Kwon et~al.(2023)Kwon, Li, Zhuang, Sheng, Zheng, Yu, Gonzalez, Zhang, and Stoica]{kwon2023efficient}
Woosuk Kwon, Zhuohan Li, Siyuan Zhuang, Ying Sheng, Lianmin Zheng, Cody~Hao Yu, Joseph Gonzalez, Hao Zhang, and Ion Stoica.
\newblock Efficient memory management for large language model serving with pagedattention.
\newblock In \emph{Proceedings of the 29th symposium on operating systems principles}, pages 611--626, 2023.

\bibitem[Lee et~al.(2026)Lee, Choi, Ko, Kim, and Kim]{lee2026videosearch}
Seohyun Lee, Seoung Choi, Dohwan Ko, Jongha Kim, and Hyunwoo~J Kim.
\newblock Videosearch-r1: Iterative video retrieval and reasoning via soft query refinement.
\newblock In \emph{European Conference on Computer Vision}, pages 648--664. Springer, 2026.

\bibitem[Lei et~al.(2020)Lei, Yu, Berg, and Bansal]{lei2020tvr}
Jie Lei, Licheng Yu, Tamara~L Berg, and Mohit Bansal.
\newblock Tvr: A large-scale dataset for video-subtitle moment retrieval.
\newblock In \emph{European Conference on Computer Vision}, pages 447--463. Springer, 2020.

\bibitem[Lei et~al.(2021)Lei, Berg, and Bansal]{lei2021mtvr}
Jie Lei, Tamara Berg, and Mohit Bansal.
\newblock mtvr: Multilingual moment retrieval in videos.
\newblock In \emph{Proceedings of the 59th Annual Meeting of the Association for Computational Linguistics and the 11th International Joint Conference on Natural Language Processing (Volume 2: Short Papers)}, pages 726--734, 2021.

\bibitem[Li et~al.(2020)Li, Chen, Cheng, Gan, Yu, and Liu]{li2020hero}
Linjie Li, Yen-Chun Chen, Yu Cheng, Zhe Gan, Licheng Yu, and Jingjing Liu.
\newblock Hero: Hierarchical encoder for video+ language omni-representation pre-training.
\newblock In \emph{Proceedings of the 2020 conference on empirical methods in natural language processing (EMNLP)}, pages 2046--2065, 2020.

\bibitem[Li et~al.(2023)Li, Shi, Leng, Zhou, Zheng, and Zhang]{li2023learning}
Mingchao Li, Xiaoming Shi, Haitao Leng, Wei Zhou, Hai-Tao Zheng, and Kuncai Zhang.
\newblock Learning semantic alignment with global modality reconstruction for video-language pre-training towards retrieval.
\newblock In \emph{Proceedings of the AAAI Conference on Artificial Intelligence}, pages 1377--1385, 2023.

\bibitem[Li et~al.(2026)Li, Zhang, Long, Chen, Song, Bai, Yang, Xie, Yang, Liu, et~al.]{li2026qwen3}
Mingxin Li, Yanzhao Zhang, Dingkun Long, Keqin Chen, Sibo Song, Shuai Bai, Zhibo Yang, Pengjun Xie, An Yang, Dayiheng Liu, et~al.
\newblock Qwen3-vl-embedding and qwen3-vl-reranker: A unified framework for state-of-the-art multimodal retrieval and ranking.
\newblock \emph{arXiv preprint arXiv:2601.04720}, 2026.

\bibitem[Liu et~al.(2025)Liu, Wang, Zhang, Wang, Xu, Wang, Yan, Zhang, and Chen]{liu2025towards}
Jingren Liu, Yun Wang, Long Zhang, Yiheng Wang, Shuning Xu, Ling Wang, Jiaqi Yan, Dell Zhang, and Xiangyu Chen.
\newblock Towards training-free long video understanding: methods, benchmarks, and open challenges.
\newblock \emph{Vicinagearth}, 2\penalty0 (1):\penalty0 6, 2025.

\bibitem[Ma and Ngo(2022)]{ma2022interactive}
Zhixin Ma and Chong~Wah Ngo.
\newblock Interactive video corpus moment retrieval using reinforcement learning.
\newblock In \emph{Proceedings of the 30th ACM International Conference on Multimedia}, pages 296--306, 2022.

\bibitem[Maeoki et~al.(2021)Maeoki, Mukuta, and Harada]{maeoki2021video}
Sho Maeoki, Yusuke Mukuta, and Tatsuya Harada.
\newblock Video moment retrieval with text query considering many-to-many correspondence using potentially relevant pair.
\newblock \emph{arXiv preprint arXiv:2106.13566}, 2021.

\bibitem[Meng et~al.(2025)Meng, Liu, Xin, Guo, and Zou]{meng2025needle}
Lingwen Meng, Fangyuan Liu, Mingyong Xin, Siqi Guo, and Fu Zou.
\newblock Needle in a haystack: Coarse-to-fine alignment network for moment retrieval from large-scale video collections.
\newblock \emph{Plos one}, 20\penalty0 (5):\penalty0 e0320661, 2025.

\bibitem[Otani et~al.(2020)Otani, Nakashima, Rahtu, and Heikkil{\"a}]{otani2020uncovering}
Mayu Otani, Yuta Nakashima, Esa Rahtu, and Janne Heikkil{\"a}.
\newblock Uncovering hidden challenges in query-based video moment retrieval.
\newblock \emph{arXiv preprint arXiv:2009.00325}, 2020.

\bibitem[Pan et~al.(2025)Pan, Zhang, Kampffmeyer, and Zhao]{pan2025refcap}
Yi Pan, Yujia Zhang, Michael Kampffmeyer, and Xiaoguang Zhao.
\newblock Refcap: Zero-shot video corpus moment retrieval based on refined dense video captioning.
\newblock In \emph{ICASSP 2025-2025 IEEE International Conference on Acoustics, Speech and Signal Processing (ICASSP)}, pages 1--5. IEEE, 2025.

\bibitem[Qin et~al.(2024)Qin, Jagerman, Hui, Zhuang, Wu, Yan, Shen, Liu, Liu, Metzler, et~al.]{qin2024large}
Zhen Qin, Rolf Jagerman, Kai Hui, Honglei Zhuang, Junru Wu, Le Yan, Jiaming Shen, Tianqi Liu, Jialu Liu, Donald Metzler, et~al.
\newblock Large language models are effective text rankers with pairwise ranking prompting.
\newblock In \emph{Findings of the Association for Computational Linguistics: NAACL 2024}, pages 1504--1518, 2024.

\bibitem[Radovanovic et~al.(2010)Radovanovic, Nanopoulos, and Ivanovic]{radovanovic2010hubs}
Milos Radovanovic, Alexandros Nanopoulos, and Mirjana Ivanovic.
\newblock Hubs in space: Popular nearest neighbors in high-dimensional data.
\newblock \emph{Journal of machine learning research}, 11\penalty0 (sept):\penalty0 2487--2531, 2010.

\bibitem[Ren et~al.(2024)Ren, Yao, Li, Sun, and Hou]{ren2024timechat}
Shuhuai Ren, Linli Yao, Shicheng Li, Xu Sun, and Lu Hou.
\newblock Timechat: A time-sensitive multimodal large language model for long video understanding.
\newblock In \emph{2024 IEEE/CVF Conference on Computer Vision and Pattern Recognition (CVPR)}, pages 14313--14323. IEEE, 2024.

\bibitem[Schroeder et~al.(2025)Schroeder, Desai, Cuadron, Chu, Liu, Zhao, Krusche, Kemper, Zaharia, and Gonzalez]{schroeder2025vcache}
Luis~Gaspar Schroeder, Aditya Desai, Alejandro Cuadron, Kyle Chu, Shu Liu, Mark Zhao, Stephan Krusche, Alfons Kemper, Matei Zaharia, and Joseph~E Gonzalez.
\newblock vcache: Verified semantic prompt caching.
\newblock \emph{arXiv preprint arXiv:2502.03771}, 2025.

\bibitem[Shinn et~al.(2023)Shinn, Cassano, Gopinath, Narasimhan, and Yao]{shinn2023reflexion}
Noah Shinn, Federico Cassano, Ashwin Gopinath, Karthik Narasimhan, and Shunyu Yao.
\newblock Reflexion: Language agents with verbal reinforcement learning.
\newblock \emph{Advances in neural information processing systems}, 36:\penalty0 8634--8652, 2023.

\bibitem[Sigurdsson et~al.(2016)Sigurdsson, Varol, Wang, Farhadi, Laptev, and Gupta]{sigurdsson2016hollywood}
Gunnar~A Sigurdsson, G{\"u}l Varol, Xiaolong Wang, Ali Farhadi, Ivan Laptev, and Abhinav Gupta.
\newblock Hollywood in homes: Crowdsourcing data collection for activity understanding.
\newblock In \emph{European conference on computer vision}, pages 510--526. Springer, 2016.

\bibitem[Sun et~al.(2023)Sun, Yan, Ma, Wang, Ren, Chen, Yin, and Ren]{sun2023chatgpt}
Weiwei Sun, Lingyong Yan, Xinyu Ma, Shuaiqiang Wang, Pengjie Ren, Zhumin Chen, Dawei Yin, and Zhaochun Ren.
\newblock Is chatgpt good at search? investigating large language models as re-ranking agents.
\newblock In \emph{Proceedings of the 2023 conference on empirical methods in natural language processing}, pages 14918--14937, 2023.

\bibitem[Sun et~al.(2026)Sun, Liu, Li, Wu, Zong, Zhang, and Liang]{sun2026genspan}
Yunzhuo Sun, Xinyue Liu, Yanyang Li, Nanding Wu, Linlin Zong, Xianchao Zhang, and Wenxin Liang.
\newblock Genspan: Generation-calibrated motion span priors for multi-verb video corpus moment retrieval.
\newblock \emph{arXiv preprint arXiv:2603.22121}, 2026.

\bibitem[Suzgun et~al.(2026)Suzgun, Yuksekgonul, Bianchi, Jurafsky, and Zou]{suzgun2026dynamic}
Mirac Suzgun, Mert Yuksekgonul, Federico Bianchi, Dan Jurafsky, and James Zou.
\newblock Dynamic cheatsheet: Test-time learning with adaptive memory.
\newblock In \emph{Proceedings of the 19th Conference of the European Chapter of the Association for Computational Linguistics (Volume 1: Long Papers)}, pages 7080--7106, 2026.

\bibitem[Wang et~al.(2024)Wang, Zhang, Zohar, and Yeung-Levy]{wang2024videoagent}
Xiaohan Wang, Yuhui Zhang, Orr Zohar, and Serena Yeung-Levy.
\newblock Videoagent: Long-form video understanding with large language model as agent.
\newblock In \emph{European Conference on Computer Vision}, pages 58--76. Springer, 2024.

\bibitem[Wang et~al.(2026)Wang, Wang, Xu, Du, Lin, Xiao, Yue, Ju, Zhang, Yang, et~al.]{wang2026time}
Ye Wang, Ziheng Wang, Boshen Xu, Yang Du, Kejun Lin, Zihan Xiao, Zihao Yue, Jianzhong Ju, Liang Zhang, Dingyi Yang, et~al.
\newblock Time-r1: Post-training large vision language model for temporal video grounding.
\newblock \emph{Advances in Neural Information Processing Systems}, 38:\penalty0 83330--83364, 2026.

\bibitem[Wang et~al.(2025)Wang, Yu, Stengel-Eskin, Yoon, Cheng, Bertasius, and Bansal]{wang2025videotree}
Ziyang Wang, Shoubin Yu, Elias Stengel-Eskin, Jaehong Yoon, Feng Cheng, Gedas Bertasius, and Mohit Bansal.
\newblock Videotree: Adaptive tree-based video representation for llm reasoning on long videos.
\newblock In \emph{2025 IEEE/CVF Conference on Computer Vision and Pattern Recognition (CVPR)}, pages 3272--3282. IEEE, 2025.

\bibitem[Xu et~al.(2026)Xu, Ye, Wu, Li, Wang, Xia, Lu, Maneriker, Du, Li, et~al.]{xu2026towards}
Chenwei Xu, Zhen Ye, Shang Wu, Weijian Li, Zihan Wang, Zhuofan Xia, Lie Lu, Pranav Maneriker, Fan Du, Manling Li, et~al.
\newblock Towards sparse video understanding and reasoning.
\newblock \emph{arXiv preprint arXiv:2602.13602}, 2026.

\bibitem[Yoon et~al.(2022)Yoon, Hong, Yoon, Kim, Kim, Yoon, and Yoo]{yoon2022selective}
Sunjae Yoon, Ji~Woo Hong, Eunseop Yoon, Dahyun Kim, Junyeong Kim, Hee~Suk Yoon, and Chang~D Yoo.
\newblock Selective query-guided debiasing for video corpus moment retrieval.
\newblock In \emph{European Conference on Computer Vision}, pages 185--200. Springer, 2022.

\bibitem[Yuan et~al.(2021)Yuan, Lan, Wang, Chen, Wang, and Zhu]{yuan2021closer}
Yitian Yuan, Xiaohan Lan, Xin Wang, Long Chen, Zhi Wang, and Wenwu Zhu.
\newblock A closer look at temporal sentence grounding in videos: Dataset and metric.
\newblock In \emph{Proceedings of the 2nd international workshop on human-centric multimedia analysis}, pages 13--21, 2021.

\bibitem[Zhang et~al.(2020{\natexlab{a}})Zhang, Hu, Lee, Zhao, Chammas, Jain, Ie, and Sha]{zhang2020hierarchical}
Bowen Zhang, Hexiang Hu, Joonseok Lee, Ming Zhao, Sheide Chammas, Vihan Jain, Eugene Ie, and Fei Sha.
\newblock A hierarchical multi-modal encoder for moment localization in video corpus.
\newblock \emph{arXiv preprint arXiv:2011.09046}, 2020{\natexlab{a}}.

\bibitem[Zhang et~al.(2024)Zhang, Lu, Islam, Wang, Yu, Bansal, and Bertasius]{zhang2024simple}
Ce Zhang, Taixi Lu, Md~Mohaiminul Islam, Ziyang Wang, Shoubin Yu, Mohit Bansal, and Gedas Bertasius.
\newblock A simple llm framework for long-range video question-answering.
\newblock In \emph{Proceedings of the 2024 Conference on Empirical Methods in Natural Language Processing}, pages 21715--21737, 2024.

\bibitem[Zhang et~al.(2021)Zhang, Sun, Jing, Nan, Zhen, Zhou, and Goh]{zhang2021video}
Hao Zhang, Aixin Sun, Wei Jing, Guoshun Nan, Liangli Zhen, Joey~Tianyi Zhou, and Rick Siow~Mong Goh.
\newblock Video corpus moment retrieval with contrastive learning.
\newblock In \emph{Proceedings of the 44th International ACM SIGIR Conference on Research and Development in Information Retrieval}, pages 685--695, 2021.

\bibitem[Zhang et~al.(2026)Zhang, Wang, Ge, Ge, Li, and Wang]{zhang2026timelens}
Jun Zhang, Teng Wang, Yuying Ge, Yixiao Ge, Xinhao Li, and Limin Wang.
\newblock Timelens: Rethinking video temporal grounding with multimodal llms.
\newblock In \emph{Proceedings of the IEEE/CVF Conference on Computer Vision and Pattern Recognition}, pages 10419--10429, 2026.

\bibitem[Zhang et~al.(2025)Zhang, Song, Duan, Wang, Chang, and Yang]{zhang2025video}
Long Zhang, Peipei Song, Zhangling Duan, Shuo Wang, Xiaojun Chang, and Xun Yang.
\newblock Video corpus moment retrieval with query-specific context learning and progressive localization.
\newblock \emph{IEEE Transactions on Circuits and Systems for Video Technology}, 35\penalty0 (6):\penalty0 5659--5670, 2025.

\bibitem[Zhang et~al.(2020{\natexlab{b}})Zhang, Peng, Fu, and Luo]{zhang2020learning}
Songyang Zhang, Houwen Peng, Jianlong Fu, and Jiebo Luo.
\newblock Learning 2d temporal adjacent networks for moment localization with natural language.
\newblock In \emph{Proceedings of the AAAI conference on artificial intelligence}, pages 12870--12877, 2020{\natexlab{b}}.

\bibitem[Zhao et~al.(2026)Zhao, Li, Zhang, Wang, and Hua]{zhao2026training}
Jialong Zhao, Huafeng Li, Yafei Zhang, Wen Wang, and Changchun Hua.
\newblock Training-free video corpus moment retrieval via synergistic collaboration and adaptive calibration.
\newblock \emph{IEEE Transactions on Image Processing}, 2026.

\bibitem[Zheng et~al.(2024{\natexlab{a}})Zheng, Yin, Xie, Sun, Huang, Yu, Cao, Kozyrakis, Stoica, Gonzalez, et~al.]{zheng2024sglang}
Lianmin Zheng, Liangsheng Yin, Zhiqiang Xie, Chuyue Sun, Jeff Huang, Cody~H Yu, Shiyi Cao, Christos Kozyrakis, Ion Stoica, Joseph~E Gonzalez, et~al.
\newblock Sglang: Efficient execution of structured language model programs.
\newblock \emph{Advances in neural information processing systems}, 37:\penalty0 62557--62583, 2024{\natexlab{a}}.

\bibitem[Zheng et~al.(2024{\natexlab{b}})Zheng, Cai, Chen, Peng, and Liu]{zheng2024training}
Minghang Zheng, Xinhao Cai, Qingchao Chen, Yuxin Peng, and Yang Liu.
\newblock Training-free video temporal grounding using large-scale pre-trained models.
\newblock In \emph{European Conference on Computer Vision}, pages 20--37. Springer, 2024{\natexlab{b}}.

\end{thebibliography}
}

\makeatletter
\let\supp@table\table \let\endsupp@table\endtable
\renewenvironment{table}[1][]{\supp@table[!htb]}{\endsupp@table}
\renewenvironment{table*}[1][]{\supp@table[!htb]}{\endsupp@table}
\let\supp@figure\figure \let\endsupp@figure\endfigure
\renewenvironment{figure}[1][]{\supp@figure[!htb]}{\endsupp@figure}
\renewenvironment{figure*}[1][]{\supp@figure[!htb]}{\endsupp@figure}
\makeatother

\clearpage
\onecolumn
\appendix
\renewcommand{\thesection}{\Alph{section}}
\crefalias{section}{appendix}
\crefalias{subsection}{appendix}

\begin{center}
{\Large \textbf{\thetitle}}\\
\vspace{0.5em}
Supplementary Material \\
\vspace{1.0em}
\end{center}

\setlength{\parskip}{0.3em}

{\large\bfseries
\begin{itemize}
  \item \Cref{sec:supp_notation}: \nameref*{sec:supp_notation}
  \item \Cref{sec:supp_method}: \nameref*{sec:supp_method}
  \item \Cref{sec:supp_setup}: \nameref*{sec:supp_setup}
  \item \Cref{sec:supp_results}: \nameref*{sec:supp_results}
  \item \Cref{sec:supp_ablations}: \nameref*{sec:supp_ablations}
  \item \Cref{sec:supp_stream}: \nameref*{sec:supp_stream}
  \item \Cref{sec:supp_analysis}: \nameref*{sec:supp_analysis}
  \item \Cref{sec:supp_limitations}: \nameref*{sec:supp_limitations}
  \item \Cref{sec:supp_prompts}: \nameref*{sec:supp_prompts}
\end{itemize}
}
\clearpage

\section{Notation}
\label{sec:supp_notation}
\begin{table}[h]
\centering
\caption{\textbf{Notation.} The last row lists the meanings of $k$ in the main text and appendix.}
\label{tab:notation}
\small
\renewcommand{\arraystretch}{0.9}
\begin{tabular}{ll}
\toprule
Notation & Description \\
\midrule
$[a,b]$ & Temporal interval from time $a$ to time $b$ \\
$\cosine(\cdot,\cdot)$ & Cosine similarity of two embeddings \\
$\iou(\cdot,\cdot)$ & Temporal intersection over union of two intervals \\
$\mathrm{VLM}(\cdot)$ & Output of the VLM for the given text and frames \\
$\mathrm{LLM}(\cdot)$ & Output of the LLM for the given text \\
\midrule
$\mathcal{V}$ & Corpus of videos $\{v_i\}_{i=1}^{N}$ \\
$N$ & Number of videos in the corpus \\
$v_i$ & $i$-th video in the corpus \\
$v$ & Video in the corpus \\
$T_v$ & Duration of video $v$ in frames \\
$t$ & Time in a video \\
$x_v(t)$ & Frame of video $v$ at time $t$ \\
$q$ & Text query \\
$v^\star$ & Ground-truth video of query $q$ \\
$t_s^\star$ & Start time of the ground-truth interval in $v^\star$ \\
$t_e^\star$ & End time of the ground-truth interval in $v^\star$ \\
$\pi$ & Final ranking of the corpus, $(\pi_1,\ldots,\pi_N)$ \\
$\pi_i$ & Video at rank $i$ in the final ranking \\
$\hat t_s$ & Start time of the predicted interval in $\pi_1$ (\cref{eq:localize}) \\
$\hat t_e$ & End time of the predicted interval in $\pi_1$ (\cref{eq:localize}) \\
$\theta$ & IoU threshold of R@1 (0.5 and 0.7) \\
\midrule
$q^{\text{emb}}$ & Embedding of query $q$ \\
$x_v^{\text{emb}}(t)$ & Embedding of frame $x_v(t)$ (frames sampled at 2 fps) \\
$s(q,v)$ & Video score, the highest similarity between $q$ and the frames of $v$ (\cref{eq:score}) \\
$\mathcal{Q}$ & Query bank (all input queries in the batch setting) \\
$q'$ & Query in the query bank \\
$s_{(j)}(v)$ & $j$-th highest similarity score of video $v$ across all queries in $\mathcal{Q}$ \\
$K_b$ & Number of top similarity scores averaged into the bias (20) \\
$b(v)$ & Video-specific bias, the average of the top $K_b$ scores of video $v$ \\
$\alpha$ & Strength of the penalty (0.75) \\
$\tilde s(q,v)$ & Adjusted score $s(q,v)-\alpha b(v)$ (\cref{eq:bank}) \\
$K$ & Number of candidate videos (10) \\
$\mathcal{C}(q)$ & Candidate list of query $q$, the top $K$ videos by $\tilde s(q,v)$, $(c_1,\ldots,c_K)$ \\
$c_i$ & Video at rank $i$ in the candidate list \\
\midrule
$t_j$ & $j$-th frame index sampled for a VLM call, evenly spaced over the video \\
$\mathcal{N}(v)$ & Notes of video $v$ (\cref{eq:notes}) \\
$n_v$ & Number of frames of video $v$ in its notes call (at most 64, at most 2 fps) \\
$o_i$ & Video at rank $i$ in the notes-based ranking (\cref{eq:rank}) \\
$\rho_{\text{notes}}(v)$ & Notes-based rank of video $v$ \\
$\rho_{\text{search}}(v)$ & Embedding-based rank of video $v$ \\
$r(v)$ & Combined score of video $v$ by reciprocal rank fusion (\cref{eq:fusion}) \\
\midrule
$u$ & Top-ranked video $\pi_1$ \\
$m$ & Number of frames in a localization call (at most 120, at most 4 fps) \\
$\Delta$ & Frame spacing $t_2-t_1$; the second run reads denser frames from $[\hat t_s-\Delta,\hat t_e+\Delta]$ \\
$w$ & Margin added to each side of $[\hat t_s,\hat t_e]$ (2\,s, \cref{sec:vcmr_results}) \\
\midrule
$k$ & \begin{tabular}[t]{@{}ll@{}}
\cref{sec:related} & Number of nearest queries in NNN~\cite{chowdhury2024nearest} \\
\cref{sec:method} & Number of top-ranked videos in VR R@$k$ (1 and 10) \\
\cref{eq:fusion} & RRF constant (60) \\
\cref{tab:supp_rounds} & Number of top-ranked videos in the upper bound (3 and 10) \\
\cref{tab:supp_calibration} & Number of labeled development queries used to select a margin (5, 10, 20 and 50)
\end{tabular} \\
\bottomrule
\end{tabular}
\end{table}

\section{Methodology}
\label{sec:supp_method}
Each candidate video needs one set of notes, even if several queries retrieve it. Once the notes are available, we rank candidates independently for each query (\cref{alg:relay}). For localization, queries that select the same video take turns using its cached frames. Different videos can be processed in parallel.

\begin{algorithm}[h]
\caption{\method over a batch of queries}
\label{alg:relay}
\small
\begin{algorithmic}[1]
\Require queries $\mathcal{Q}$, corpus $\mathcal{V}$ with frame embeddings
\State $s(q,v)\gets\max_t\cosine(q^{\text{emb}},x_v^{\text{emb}}(t))$ for all $q\in\mathcal{Q}$, $v\in\mathcal{V}$ \Comment{search}
\State $b(v)\gets$ mean of the $K_b$ highest $s(q',v)$, $q'\in\mathcal{Q}$;\ $\tilde s(q,v)\gets s(q,v)-\alpha b(v)$ \Comment{query bank}
\State $\mathcal{C}(q)\gets$ top-$K$ videos by $\tilde s(q,\cdot)$, for every $q$
\For{$v\in\bigcup_q\mathcal{C}(q)$, in parallel} \Comment{ranking}
  \State $\mathcal{N}(v)\gets\mathrm{VLM}(n_v$ frames evenly spaced over $v)$ \Comment{video notes}
\EndFor
\For{$q\in\mathcal{Q}$, in parallel}
  \State $(o_1,\ldots,o_K)\gets\mathrm{LLM}(q,\mathcal{N}(c_1),\ldots,\mathcal{N}(c_K))$ \Comment{notes-based ranking}
  \State $\pi_1,\ldots,\pi_K\gets\mathcal{C}(q)$ sorted by $r(v)$, ties by $\rho_{\text{search}}$; then the other videos by $\tilde s$ \Comment{fusion}
\EndFor
\For{$u\in\{\pi_1(q):q\in\mathcal{Q}\}$, in parallel} \Comment{localization}
  \State form a shared prefix from $m$ timestamped frames of $u$ and the initial VLM response giving its span
  \For{$q$ with $\pi_1(q)=u$, one at a time} \Comment{video prompt caching}
    \State $[\hat t_s,\hat t_e]\gets\mathrm{VLM}(\text{shared prefix},q)$
    \State if the sampling rate is at most 2 fps, update $[\hat t_s,\hat t_e]$ with a separate call on the padded window (\cref{sec:supp_localize})
    \State output $(\pi(q),[\hat t_s-w,\ \hat t_e+w])$, clipped to the video \Comment{margin}
  \EndFor
\EndFor
\end{algorithmic}
\end{algorithm}

\subsection{Search}
\label{sec:supp_search}
We encode each query with Qwen3-VL-Embedding-2B~\cite{li2026qwen3} using the instruction ``Find a video that contains the following visual content.''
We sample video frames at 2\,fps and score each video by its highest frame similarity (\cref{eq:score}).
We chose $K_b=20$ and $\alpha=0.75$ by maximizing search VR R@1 on 2,693 Charades-FIG training queries (\cref{sec:supp_data}).
At test time we apply these settings to the split's query bank and retain the ten videos with the highest adjusted scores (\cref{eq:bank}).

\subsection{Ranking}
\label{sec:supp_rank}
\noindent \textbf{Video Notes.}
For each candidate video we sample frames uniformly at 2\,fps, reducing the rate when needed to stay within 64 frames. The VLM describes the setting, people and objects, then lists the actions in time order with their start and end times (\cref{fig:notes_example}).

\begin{figure}[h]
\centering
\begin{promptfloat}[Notes for the top-ranked video in \cref{fig:framework} (10.6\,s, 22 frames, 4,564 input tokens, 386 output tokens)]{palegreen}
\begin{verbatim}
Place: A home kitchen with white walls and a light countertop. Brown wooden upper cabinets with
  dark round knobs line the back wall. A white range hood is above a stove at left; a microwave,
  small containers and a bowl sit on the counter. A sink and curved faucet are at far right.
People: A young adult man with light skin and curly dark brown hair, wearing a red long-sleeved
  shirt and dark trousers.
Objects: Brown wooden cabinet doors with dark knobs; the narrow cabinet door beside the large
  rightmost cabinet opens outward.; A red-and-white plastic food package taken from the cabinet;
  its exact contents and label are unclear.; A stove with dark burners and metallic cookware, a
  white range hood, a dark microwave, small counter containers, a bowl, and a sink with a curved
  faucet.
Timeline:
1.0-2.0 s: The man enters from the left and walks right toward the upper cabinets.
2.0-2.5 s: Facing the cabinets, he reaches with his right hand and pulls open the narrow wooden
  cabinet door beside the large rightmost cabinet, swinging it outward.
2.5-3.5 s: He reaches into the open cabinet with his right hand and takes out a red-and-white
  plastic package.
3.5-4.5 s: He lowers the package and holds it in both hands, looking down at it with a smile.
4.5-5.0 s: Still holding the package in his left hand, he pushes the cabinet door shut with his
  right hand.
5.0-5.5 s: Holding the package, he turns and walks left, exiting the frame.
\end{verbatim}
\end{promptfloat}
\vspace{-0.1in}
\caption{\textbf{Video notes supplied to the ranker} for the cabinet-opening example in \cref{fig:framework}.
The timeline distinguishes opening the cabinet, removing a package and closing the door.}
\label{fig:notes_example}
\end{figure}

\noindent \textbf{Notes-Based Ranking.}
We number the ten candidates in search order and give the LLM their notes and the query.
It returns the candidate numbers in order of relevance.
We fuse this ranking with the search ranking using reciprocal rank fusion ($k=60$, \cref{eq:fusion}) and break ties by search rank.

\subsection{Localization}
\label{sec:supp_localize}
We sample the whole video at up to 4\,fps, with a budget of 120 frames.
For videos of 60\,s or longer, this gives at most 2\,fps.
We then take a second look at the predicted interval, adding one first-pass frame spacing $\Delta$ on either side and clipping the window to the video.
A separate call samples this window with the same frame budget and rate cap and returns the final interval.

\noindent \textbf{Video Prompt Caching.}
For each video we send the timestamped frames and ask the model for the start and end of the whole video.
The initial request and its answer form a shared prefix. Each localization request repeats that prefix and appends the current query.
We derive the prompt cache key from the initial video message and reuse it across requests for that video.
Earlier queries and their predictions never enter the prefix.

\FloatBarrier
\section{Experimental Setup}
\label{sec:supp_setup}

\subsection{Benchmarks and Metrics}
\label{sec:supp_data}
\begin{table}[h]
\centering
\caption{\textbf{Benchmarks.} Splits of the VCMR evaluation and the three development sets.
Charades-FIG, corpus: the training videos among which its development set is searched, with all their queries.
ActivityNet Captions annotates its validation videos twice: val is the first annotation, and the ActivityNet-FIG test split rewrites the second.
Video and Moment: mean length in seconds; Words: mean query length.
$^a$Annotated first 30\,s; the video files last 50.0\,s (DiDeMo-FIG) and 50.1\,s (DiDeMo) on average.}
\label{tab:supp_benchmarks}
\small
\setlength{\tabcolsep}{4pt}
\begin{tabular}{llllrrrrr}
\toprule
Task & Dataset & Query type & Split & Videos & Queries & Video (s) & Moment (s) & Words \\
\midrule
VCMR, VR & Charades-FIG~\cite{chen2024verified} & fine-grained & test & 1,334 & 3,720 & 29.6 & 8.0 & 15.4 \\
& DiDeMo-FIG~\cite{chen2024verified} & fine-grained & test & 1,002 & 4,013 & 29.1$^a$ & 6.5 & 16.3 \\
& ActivityNet-FIG~\cite{chen2024verified} & fine-grained & test & 4,866 & 16,966 & 118.1 & 40.2 & 25.8 \\
VCMR & Charades-STA~\cite{gao2017tall} & original & test & 1,334 & 3,720 & 29.6 & 8.0 & 6.2 \\
& DiDeMo~\cite{anne2017localizing} & original & test & 1,004 & 4,021 & 29.1$^a$ & 6.5 & 7.5 \\
& ActivityNet Captions~\cite{krishna2017dense} & original & val & 4,917 & 17,505 & 118.2 & 37.7 & 13.6 \\
& TVR~\cite{lei2020tvr} & original & val & 2,179 & 10,895 & 75.7 & 9.2 & 12.2 \\
\midrule
Development & Charades-FIG, corpus & fine-grained & train & 1,334 & 3,180 & 30.6 & 8.2 & 15.3 \\
& Charades-FIG & fine-grained & train & 210 & 487 & 30.1 & 8.2 & 15.5 \\
& DiDeMo-FIG & fine-grained & val & 416 & 500 & 28.8$^a$ & 6.1 & 16.4 \\
& ActivityNet-FIG & fine-grained & val & 475 & 500 & 129.6 & 39.9 & 26.2 \\
\bottomrule
\end{tabular}
\end{table}

\noindent \textbf{VERIFIED Benchmarks.}
Charades-FIG, DiDeMo-FIG and ActivityNet-FIG~\cite{chen2024verified} pair fine-grained queries with the videos and moment annotations of Charades-STA~\cite{gao2017tall}, DiDeMo~\cite{anne2017localizing} and ActivityNet Captions~\cite{krishna2017dense}, respectively (\cref{tab:supp_benchmarks}).
DiDeMo annotates only the first 30\,s of each video, in 5-second segments.
DiDeMo-FIG uses the span most frequently marked by the four annotators.

\noindent \textbf{Original Queries.}
For VCMR with the original queries, we use the Charades-STA and DiDeMo test splits and the first validation annotation of ActivityNet Captions.
Trained VCMR models report their results on these splits~\cite{zhang2021video,yoon2022selective}.

\noindent \textbf{TVR.}
TVR~\cite{lei2020tvr} contains subtitled clips from six TV shows.
We use the released 3\,fps frames and, for \method\ w/ subtitles, the released subtitles.
Of the 10,895 validation queries, TVR labels 8,097 as answerable from video alone.

\noindent \textbf{Development Sets.}
For Charades-FIG, we sample 1,334 training videos to match the test corpus size.
We evaluate on 487 queries from 210 of those videos, searching all 1,334.
The remaining 2,693 queries are used to tune the query-bank parameters (\cref{sec:supp_search}).
For DiDeMo-FIG, we sample 500 validation queries and search their 416 videos.
For ActivityNet-FIG, we sample 500 queries from the first validation annotation and search the 4,662 available videos in the test corpus.

\noindent \textbf{Metrics.}
In addition to the main thresholds of 0.5 and 0.7, we report R@1 at IoU 0.3 where indicated.
We clip each annotated moment to the length of its video before computing the IoU.
All 16,966 ActivityNet-FIG test queries enter the recall denominator.
Queries referring to the 204 unavailable videos count as misses.

\subsection{Implementation Details}
\label{sec:supp_impl}
We use gpt-6-luna for both visual and text-only calls, with reasoning effort set to ``none''.
We call the model through the Codex CLI and, on TVR, through the OpenAI Chat Completions API.
With subtitles on TVR, 197 of the 2,598 queries whose top-ranked video is the annotated one are localized through the Codex CLI.
Each frame is a JPEG (quality 90, at most $448\times448$ pixels in area) preceded by its timestamp.
We use the prompts in \cref{sec:supp_prompts}, require responses to match a JSON schema and allow up to three attempts per call.
The code release includes the pipeline, experimental variants, evaluation scripts and tests.

\subsection{Margin}
\label{sec:supp_calibration}
We compare six fixed margins from 0 to 5\,s per side.
Of these, 2\,s gives the highest development VCMR R@1 at IoU 0.5 on each benchmark (\cref{tab:dev_margin}).
We use this common margin for the test results reported as ``w/ margin.'' The lower rows apply the same sweep to model and human intervals in the annotated video.
ActivityNet-FIG retains 1,284 of the 1,288 sentences with independent human annotations.
\begin{table}[t]
\centering
\caption{\textbf{Margin on the Development Sets and for Human Annotators}: R@1 (\%) at IoU 0.5 with the interval widened by $w$ seconds on each side.
\method, VCMR: on the three development sets.
Annotated video: on the 1,000 Charades-FIG and 1,284 ActivityNet-FIG test queries whose Charades-STA and ActivityNet Captions sentences five annotators of \citet{otani2020uncovering} re-marked.
\method\ and each annotator localize that sentence in the annotated video, and the annotators' R@1 is averaged over the five.
Best per row in bold.}
\label{tab:dev_margin}
\vspace{-0.08in}
\small
\setlength{\tabcolsep}{4pt}
\begin{tabular}{lcccccc}
\toprule
& \multicolumn{6}{c}{Margin $w$ (s)} \\
\cmidrule(lr){2-7}
& 0 & $+0.5$ & $+1$ & $+2$ & $+3$ & $+5$ \\
\midrule
\multicolumn{7}{l}{\method, VCMR} \\
Charades-FIG & 9.7 & 13.6 & 15.4 & \textbf{19.3} & 18.5 & 16.2 \\
DiDeMo-FIG & 15.0 & 21.8 & 30.0 & \textbf{35.2} & 34.8 & 17.2 \\
ActivityNet-FIG & 23.6 & 24.4 & 24.4 & \textbf{26.6} & 26.4 & 26.0 \\
\midrule
\multicolumn{7}{l}{\method, annotated video} \\
Charades-FIG & 23.4 & 33.6 & 51.0 & 65.6 & \textbf{66.8} & 58.9 \\
ActivityNet-FIG & 30.1 & 31.5 & 32.4 & 33.7 & 34.7 & \textbf{35.6} \\
\midrule
\multicolumn{7}{l}{Human annotator, annotated video} \\
Charades-FIG & 44.0 & 58.8 & 68.4 & \textbf{73.7} & 72.8 & 57.8 \\
ActivityNet-FIG & 35.1 & 36.2 & 36.6 & 37.6 & \textbf{37.9} & \textbf{37.9} \\
\bottomrule
\end{tabular}
\vspace{-0.1in}
\end{table}

\noindent \textbf{Selecting a Margin with Few Labels.}
In a separate experiment, we draw $k$ development queries without replacement and select a margin using only localization in the annotated video.
We search from $-5$ to $+5$\,s in steps of 0.05\,s, maximizing R@1 at IoU 0.5.
We sort tied margins and take the middle entry (the upper middle for an even count).
We then evaluate the selected margin on test queries using the retrieved video.
Across 2,000 draws of 20 queries from the same benchmark, the largest difference in mean test VCMR R@1 at IoU 0.5 from the fixed 2\,s result is 0.73 percentage points, on DiDeMo-FIG (\cref{tab:supp_calibration}).
There, mean recall is 31.7\%.
The score exceeds the published VideoSearch-R1 result in 87\% of draws there and at least 98\% on the other two splits.
Using 20 Charades-FIG training queries to select the margin for all three benchmarks keeps the mean within 1.8 points of the fixed-margin result at IoU 0.5.
At IoU 0.7, same-benchmark selection lowers mean recall by up to 1.6 points.

\begin{table*}[t]
\centering
\caption{\textbf{Selecting a margin from $k$ labeled development queries.} We choose $w$ by localization R@1 at IoU 0.5 in the annotated video, then evaluate VCMR on the test split. The two blocks use labels from the target benchmark or from Charades-FIG. For each $k$, we report the median $w$ in seconds and mean test R@1 (\%) at IoU 0.5 and 0.7 over 2,000 draws. The \emph{all} row uses the full development set (split sizes in table order: 487, 500, 500). The last two rows show the fixed $w=2$\,s result and the published VideoSearch-R1 scores~\cite{lee2026videosearch}.}
\label{tab:supp_calibration}
\vspace{-0.08in}
\small
\setlength{\tabcolsep}{4pt}
\begin{tabular}{lcccccc}
\toprule
& \multicolumn{3}{c}{Same benchmark} & \multicolumn{3}{c}{Charades-FIG} \\
\cmidrule(lr){2-4}\cmidrule(lr){5-7}
$k$ & $w$ & 0.5 & 0.7 & $w$ & 0.5 & 0.7 \\
\midrule
\multicolumn{7}{l}{Charades-FIG} \\
\quad 5 & 2.95 & 20.9 & 13.0 & -- & -- & -- \\
\quad 10 & 2.85 & 21.1 & 13.1 & -- & -- & -- \\
\quad 20 & 2.85 & 21.2 & 13.3 & -- & -- & -- \\
\quad 50 & 2.80 & 21.4 & 13.8 & -- & -- & -- \\
\quad all & 3.05 & 21.5 & 14.0 & -- & -- & -- \\
\quad $w=2$\,s & 2.00 & 21.4 & 14.8 & & & \\
\quad VideoSearch-R1 & -- & 13.4 & 8.2 & & & \\
\midrule
\multicolumn{7}{l}{DiDeMo-FIG} \\
\quad 5 & 2.65 & 30.6 & 14.2 & 2.90 & 30.3 & 13.9 \\
\quad 10 & 2.65 & 31.3 & 14.6 & 2.90 & 30.3 & 14.0 \\
\quad 20 & 2.70 & 31.7 & 14.8 & 2.85 & 30.7 & 14.2 \\
\quad 50 & 2.70 & 32.2 & 15.2 & 2.70 & 31.5 & 14.7 \\
\quad all & 2.75 & 32.6 & 15.6 & 3.05 & 32.3 & 14.9 \\
\quad $w=2$\,s & 2.00 & 32.4 & 16.4 & & & \\
\quad VideoSearch-R1 & -- & 30.2 & 19.7 & & & \\
\midrule
\multicolumn{7}{l}{ActivityNet-FIG} \\
\quad 5 & 1.90 & 24.3 & 13.6 & 2.90 & 25.0 & 13.7 \\
\quad 10 & 2.45 & 24.8 & 13.7 & 2.90 & 25.1 & 13.8 \\
\quad 20 & 2.75 & 25.0 & 13.7 & 2.85 & 25.1 & 13.8 \\
\quad 50 & 2.65 & 25.1 & 13.7 & 2.80 & 25.1 & 13.8 \\
\quad all & 2.10 & 24.9 & 13.9 & 3.05 & 25.3 & 13.8 \\
\quad $w=2$\,s & 2.00 & 24.9 & 13.9 & & & \\
\quad VideoSearch-R1 & -- & 22.3 & 12.3 & & & \\
\bottomrule
\end{tabular}
\vspace{-0.1in}
\end{table*}

\noindent \textbf{Other Forms of the Margin.}
We also extend each end of the prediction by 52\% of its duration, the proportion with the highest mean development VCMR R@1 at IoU 0.5.
On ActivityNet-FIG, the proportional and fixed margins give nearly identical test scores at this threshold (24.89\% and 24.91\%).
On Charades-FIG and DiDeMo-FIG, the proportional margin scores 4.9 and 10.3 points lower, respectively (\cref{tab:supp_margin_forms}).

\begin{table}[t]
\centering
\caption{\textbf{Other Forms of the Margin}: VCMR R@1 (\%) of \method\ on the test splits at IoU 0.5 and 0.7.
The share of 52\% has the highest mean VCMR R@1 at IoU 0.5 over the three development sets.
Two intervals: one localization call returns the action that the query names and the whole activity around it, from getting ready for the action to finishing it.}
\label{tab:supp_margin_forms}
\vspace{-0.08in}
\small
\setlength{\tabcolsep}{4pt}
\begin{tabular}{lcccccc}
\toprule
& \multicolumn{2}{c}{Charades-FIG} & \multicolumn{2}{c}{DiDeMo-FIG} & \multicolumn{2}{c}{ActivityNet-FIG} \\
\cmidrule(lr){2-3}\cmidrule(lr){4-5}\cmidrule(lr){6-7}
& 0.5 & 0.7 & 0.5 & 0.7 & 0.5 & 0.7 \\
\midrule
\method & 11.75 & 5.38 & 14.88 & 5.91 & 22.39 & 13.31 \\
\quad + 2\,s per side (\method\ w/ margin) & 21.37 & 14.84 & 32.44 & 16.37 & 24.91 & 13.92 \\
\quad + 52\% of its length per side & 16.42 & 9.17 & 22.13 & 10.04 & 24.89 & 11.60 \\
\midrule
Two intervals: the action & 10.38 & 4.78 & 13.06 & 5.28 & 22.51 & 13.31 \\
\quad + 2\,s per side & 20.19 & 14.19 & 28.38 & 12.98 & 24.75 & 13.77 \\
Two intervals: the whole activity & 14.06 & 6.99 & 16.25 & 7.40 & 25.38 & 15.17 \\
\bottomrule
\end{tabular}
\vspace{-0.1in}
\end{table}

\subsection{Baselines}
\label{sec:supp_baselines}
\noindent \textbf{Trained Models.}
VERIFIED~\cite{chen2024verified} reports HERO, XML, ReLoCLNet, CONQUER and SQuiDNet, each trained on the training split of the benchmark.
CONQUER and SQuiDNet localize within the top videos of XML.
With the original queries, \cref{tab:supp_vcmr} lists the trained models that report R@1 on these splits~\cite{escorcia2019finding,lei2020tvr,li2020hero,zhang2020hierarchical,zhang2021video,hou2021conquer,yoon2022selective,li2023learning,hou2023multi,hou2024improving,hou2026event,meng2025needle,sun2026genspan}.
CONQUER, SQuiDNet, MINUTE, PREM and EventFormer localize within their retriever's top ten videos.
\method reranks ten candidates and localizes in the first.

\noindent \textbf{Training-Free Methods.}
RefCap~\cite{pan2025refcap} captions one frame per second with BLIP or MiniGPT-v2 and groups the captions into events.
It retrieves an event by the similarity of its captions to the query.
It reports R@10 and R@100 on Charades-STA and ActivityNet Captions.
\method predicts one interval per query, so its R@10 equals its R@1.
TFVTG~\cite{zheng2024training} localizes a moment in one video.
We run its released code in \method's top-ranked video, with its Charades-STA settings and gpt-6-luna as its LLM.

\noindent \textbf{Embedding-Only Baseline.}
The embedding-only baseline ranks videos by their best frame at 2\,fps (\cref{eq:score}).
In the top-ranked video, we take the contiguous run around the best frame whose scores reach at least the midpoint between the video's minimum and maximum similarities.
The interval extends half a sampling step beyond the first and last frames in this run, clipped to the video boundaries.

\FloatBarrier
\section{Additional Retrieval Results}
\label{sec:supp_results}

\subsection{Original Queries and TVR}
\label{sec:supp_original}
On the original queries, \method without a margin reaches 1.67\%, 6.49\% and 11.01\% VCMR R@1 at IoU 0.5 on Charades-STA, DiDeMo and ActivityNet Captions, respectively.
Each exceeds the available trained-model R@1 results in \cref{tab:supp_vcmr}.
RefCap reports R@10.
At that cutoff and threshold, our single prediction with the margin scores below both RefCap variants on Charades-STA and above both on ActivityNet Captions.

On TVR, the trained baselines read the subtitles as well as the video.
With the subtitles, \method reaches 10.50\% VCMR R@1 at IoU 0.5, above XML, ReLoCLNet and HAMMER and below PREM, EventFormer and GenSpan.
The subtitles raise VR R@1 from 16.0\% to 23.9\%, since they name the characters that the queries refer to.
With frames alone, \method reaches 6.77\% VCMR R@1 at IoU 0.5, above XML (5.28\%) and below the other trained models reporting that metric.
On the 8,097 queries labeled answerable from video alone, frames-only \method reaches 8.0\%, against 5.0\% for the embedding-only baseline.
\begin{table*}[t]
\centering
\caption{\textbf{VCMR with the Original Queries and on TVR}: R@1 (\%) at IoU 0.5 and 0.7 on Charades-STA test, DiDeMo test, the first validation annotation of ActivityNet Captions, and TVR val.
Published models are trained on the benchmark's training split, except RefCap, which is training-free.
On TVR, they read the video and the subtitles, and \method reads only the frames.
W/ subtitles: the search also ranks the videos by their subtitles and fuses this ranking with the frame-based one by reciprocal rank fusion, and the notes and the localization read the subtitle lines of their frames.
``--'': not reported, or no subtitles.
\method localizes with one call per query on the first three benchmarks and with video prompt caching on TVR.
$^a$First arXiv version.
$^b$Re-trained by a later paper: XML on Charades-STA by \citet{maeoki2021video}, CONQUER on ActivityNet Captions by \citet{hou2026event}.
$^c$With automatic speech recognition (ASR) subtitles; split not stated.
$^d$Split not stated.
$^e$R@10, as RefCap reports no R@1; \method predicts one interval per query, so its R@10 equals its R@1.}
\label{tab:supp_vcmr}
\small
\setlength{\tabcolsep}{4pt}
\begin{tabular}{lcccccccc}
\toprule
& \multicolumn{2}{c}{Charades-STA} & \multicolumn{2}{c}{DiDeMo} & \multicolumn{2}{c}{ActivityNet Captions} & \multicolumn{2}{c}{TVR} \\
\cmidrule(lr){2-3}\cmidrule(lr){4-5}\cmidrule(lr){6-7}\cmidrule(lr){8-9}
Method & 0.5 & 0.7 & 0.5 & 0.7 & 0.5 & 0.7 & 0.5 & 0.7 \\
\midrule
MCN (TEF)$^a$~\cite{escorcia2019finding} & 0.13 & 0.08 & 0.88 & 0.58 & 0.12 & 0.07 & -- & -- \\
CAL (TEF)$^a$~\cite{escorcia2019finding} & 0.23 & 0.12 & 0.97 & 0.66 & 0.21 & 0.12 & -- & -- \\
XML~\cite{lei2020tvr} & 0.3$^b$ & 0.22$^b$ & 2.26 & 1.59 & -- & -- & 5.28 & 2.62 \\
HERO~\cite{li2020hero} & -- & -- & -- & 2.14 & -- & -- & -- & 5.13 \\
HAMMER~\cite{zhang2020hierarchical} & -- & -- & -- & -- & 2.94 & 1.74 & 9.19 & 5.13 \\
ReLoCLNet~\cite{zhang2021video} & -- & -- & -- & -- & 3.09 & 1.82 & 8.03 & 4.15 \\
CONQUER~\cite{hou2021conquer} & -- & -- & 3.31 & 2.79 & 2.14$^b$ & 1.33$^b$ & -- & 7.76 \\
SQuiDNet~\cite{yoon2022selective} & -- & -- & -- & 3.52$^c$ & -- & 4.43 & -- & 8.52 \\
FEEL~\cite{li2023learning} & -- & -- & 4.02$^d$ & 3.34$^d$ & -- & -- & -- & -- \\
MINUTE~\cite{hou2023multi} & -- & -- & 3.44 & 2.81 & -- & -- & -- & 10.70 \\
PREM~\cite{hou2024improving} & -- & -- & 3.51 & 2.98 & -- & -- & 16.08 & 9.01 \\
EventFormer~\cite{hou2026event} & -- & -- & 3.53 & 3.12 & 4.32 & 2.75 & 17.79 & 10.12 \\
CFAN~\cite{meng2025needle} & 0.56 & 0.38 & 2.36 & 1.54 & 1.85$^d$ & 1.12$^d$ & -- & -- \\
GenSpan~\cite{sun2026genspan} & -- & -- & -- & -- & 6.36$^d$ & 4.22$^d$ & 18.77 & 12.12 \\
\midrule
RefCap (BLIP)$^e$~\cite{pan2025refcap} & 3.95 & 2.18 & -- & -- & 10.23 & 5.00 & -- & -- \\
RefCap (MiniGPT-v2)$^e$~\cite{pan2025refcap} & 4.25 & 2.12 & -- & -- & 11.47 & 5.60 & -- & -- \\
\midrule
Embedding only & 1.05 & 0.48 & 3.98 & 1.72 & 6.12 & 3.51 & 4.27 & 2.37 \\
\method & \cellcolor{LightGray}1.67 & \cellcolor{LightGray}0.59 & \cellcolor{LightGray}6.49 & \cellcolor{LightGray}2.54 & \cellcolor{LightGray}11.01 & \cellcolor{LightGray}7.47 & \cellcolor{LightGray}6.77 & \cellcolor{LightGray}4.08 \\
\method\ w/ margin & \cellcolor{LightCyan}3.47 & \cellcolor{LightCyan}2.23 & \cellcolor{LightCyan}15.42 & \cellcolor{LightCyan}7.81 & \cellcolor{LightCyan}12.27 & \cellcolor{LightCyan}7.61 & \cellcolor{LightCyan}8.31 & \cellcolor{LightCyan}3.72 \\
\midrule
Embedding only w/ subtitles & -- & -- & -- & -- & -- & -- & 4.30 & 2.25 \\
\method\ w/ subtitles & -- & -- & -- & -- & -- & -- & \cellcolor{LightGray}10.50 & \cellcolor{LightGray}5.88 \\
\method\ w/ subtitles, w/ margin & -- & -- & -- & -- & -- & -- & \cellcolor{LightCyan}12.69 & \cellcolor{LightCyan}5.31 \\
\bottomrule
\end{tabular}
\end{table*}

\subsection{Open Model}
\label{sec:supp_open}
We replaced gpt-6-luna with Qwen3.8-27B for both visual and text-only calls on Charades-FIG.
We served it through vLLM~\cite{kwon2023efficient} with thinking disabled and temperature zero, using the same prompts, frames and search results.
Test VR R@1 rose from 32.8\% to 34.0\%.
With the 2\,s margin, VCMR R@1 fell from 21.4\% to 20.9\% at IoU 0.5 and rose from 14.8\% to 15.1\% at IoU 0.7.

\subsection{Video Length}
\label{sec:supp_length}
Reranking improves VR R@1 over query-bank search in every video-length quartile (\cref{tab:supp_length}).
Even in the longest ActivityNet-FIG quartile (179--745\,s), where the median spacing between frames used for notes is 3.23\,s, the gain is 4.4 points.
On DiDeMo-FIG and ActivityNet-FIG, VCMR R@1 with the margin is highest in the shortest quartile.
\begin{table}[t]
\centering
\caption{\textbf{Results by Video Length} on the test splits: the queries whose video has frames, in 4 bins of equal size by the length of the annotated video.
Notes: median seconds between the frames that the notes of the annotated video are written from.
VR R@1 (\%): Bank is the search with the query bank, and Gain is \method\ minus Bank.
VCMR: R@1 (\%) at IoU 0.5 of \method\ w/ margin.}
\label{tab:supp_length}
\vspace{-0.08in}
\small
\setlength{\tabcolsep}{4pt}
\begin{tabular}{lcccccc}
\toprule
& & & \multicolumn{3}{c}{VR R@1} & VCMR \\
\cmidrule(lr){4-6}
Length (s) & Queries & Notes (s) & Bank & \method & Gain & 0.5 \\
\midrule
\multicolumn{7}{l}{\textit{Charades-FIG}} \\
\quad 7--27 & 927 & 0.49 & 27.1 & 30.6 & +3.6 & 21.5 \\
\quad 27--31 & 912 & 0.50 & 30.8 & 36.7 & +5.9 & 24.8 \\
\quad 31--33 & 949 & 0.50 & 30.0 & 33.5 & +3.5 & 20.9 \\
\quad 33--72 & 932 & 0.56 & 25.5 & 30.3 & +4.7 & 18.5 \\
\multicolumn{7}{l}{\textit{DiDeMo-FIG}} \\
\quad 22--31 & 1,000 & 0.50 & 63.1 & 67.0 & +3.9 & 40.0 \\
\quad 31--41 & 1,001 & 0.54 & 63.5 & 66.7 & +3.2 & 35.1 \\
\quad 41--65 & 1,007 & 0.81 & 59.6 & 63.4 & +3.8 & 27.2 \\
\quad 65--180 & 1,005 & 1.39 & 59.7 & 61.5 & +1.8 & 27.6 \\
\multicolumn{7}{l}{\textit{ActivityNet-FIG}} \\
\quad 1--64 & 4,060 & 0.55 & 64.8 & 67.9 & +3.1 & 29.7 \\
\quad 64--122 & 4,059 & 1.43 & 66.5 & 69.7 & +3.2 & 25.9 \\
\quad 122--179 & 4,061 & 2.35 & 65.1 & 69.8 & +4.7 & 25.5 \\
\quad 179--745 & 4,063 & 3.23 & 63.9 & 68.3 & +4.4 & 23.1 \\
\bottomrule
\end{tabular}
\vspace{-0.1in}
\end{table}

\subsection{Other Intervals in the Top-Ranked Video}
\label{sec:supp_label_free}
To isolate the effect of interval selection, we keep the retrieved video fixed.
We ask for the whole activity around the named action, substitute TFVTG's prediction, or align the predicted interval to DiDeMo's segment boundaries.
Each variant improves VCMR R@1 at all three thresholds on the datasets where we tested it (\cref{tab:label_free}).
\begin{table*}[t]
\centering
\caption{\textbf{Other Intervals in the Top-Ranked Video} (test splits): VCMR R@1 (\%) at IoU 0.3, 0.5 and 0.7.
All \method\ rows use the top-ranked video of \cref{tab:vcmr} and change only the interval.
\method: the interval that the VLM predicts.
W/ the whole activity: the VLM predicts the interval of the whole activity around the action instead.
W/ TFVTG's moment: TFVTG~\cite{zheng2024training} predicts the interval, with the settings that its authors chose on Charades-STA (not run on ActivityNet-FIG).
W/ DiDeMo's segments: each boundary of the predicted interval moved to the nearest boundary of DiDeMo's 5-second segments.
W/ margin: the predicted interval widened by 2\,s on each side, as in \cref{tab:vcmr}.
VideoSearch-R1 as reported by \citet{lee2026videosearch}.
Each of these intervals raises VCMR R@1 over \method\ at every threshold.}
\label{tab:label_free}
\vspace{-0.08in}
\small
\setlength{\tabcolsep}{4pt}
\begin{tabular}{lcccccccccc}
\toprule
& & \multicolumn{3}{c}{Charades-FIG} & \multicolumn{3}{c}{DiDeMo-FIG} & \multicolumn{3}{c}{ActivityNet-FIG} \\
\cmidrule(lr){3-5}\cmidrule(lr){6-8}\cmidrule(lr){9-11}
Method & Trained & 0.3 & 0.5 & 0.7 & 0.3 & 0.5 & 0.7 & 0.3 & 0.5 & 0.7 \\
\midrule
VideoSearch-R1~\cite{lee2026videosearch} & \checkmark & 16.5 & 13.4 & 8.2 & 33.3 & 30.2 & 19.7 & 33.8 & 22.3 & 12.3 \\
\midrule
\method & -- & \cellcolor{LightGray}19.17 & \cellcolor{LightGray}11.75 & \cellcolor{LightGray}5.38 & \cellcolor{LightGray}27.73 & \cellcolor{LightGray}14.88 & \cellcolor{LightGray}5.91 & \cellcolor{LightGray}34.40 & \cellcolor{LightGray}22.39 & \cellcolor{LightGray}13.31 \\
\quad w/ the whole activity & -- & \cellcolor{LightGray}24.33 & \cellcolor{LightGray}14.06 & \cellcolor{LightGray}6.99 & \cellcolor{LightGray}31.37 & \cellcolor{LightGray}16.25 & \cellcolor{LightGray}7.40 & \cellcolor{LightGray}38.59 & \cellcolor{LightGray}25.38 & \cellcolor{LightGray}15.17 \\
\quad w/ TFVTG's moment & -- & \cellcolor{LightGray}23.15 & \cellcolor{LightGray}17.23 & \cellcolor{LightGray}8.60 & \cellcolor{LightGray}30.18 & \cellcolor{LightGray}19.89 & \cellcolor{LightGray}7.63 & -- & -- & -- \\
\quad w/ DiDeMo's segments & -- & -- & -- & -- & \cellcolor{LightGray}44.95 & \cellcolor{LightGray}41.27 & \cellcolor{LightGray}29.85 & -- & -- & -- \\
\quad w/ margin & -- & \cellcolor{LightCyan}26.59 & \cellcolor{LightCyan}21.37 & \cellcolor{LightCyan}14.84 & \cellcolor{LightCyan}43.93 & \cellcolor{LightCyan}32.44 & \cellcolor{LightCyan}16.37 & \cellcolor{LightCyan}39.13 & \cellcolor{LightCyan}24.91 & \cellcolor{LightCyan}13.92 \\
\bottomrule
\end{tabular}
\vspace{-0.12in}
\end{table*}

\noindent \textbf{DiDeMo's Segments.}
We clip predictions to the first 30\,s and round each endpoint to the nearest 5\,s boundary.
If both endpoints round to the same boundary, we use the segment containing the clipped interval's midpoint (the final segment at 30\,s).
This raises DiDeMo-FIG VCMR R@1 at IoU 0.5 from 14.9\% to 41.3\% without changing which video was retrieved.
The rounded prediction exactly matches the annotated segments on 29.5\% of queries, compared with 19.7\% for VideoSearch-R1.
Of VideoSearch-R1's released intervals, 95.1\% already have both endpoints on the grid on DiDeMo-FIG, compared with 6.1\% on Charades-FIG.
Adding 2\,s per side to its DiDeMo-FIG predictions lowers VCMR R@1 at IoU 0.5 from 30.3\% to 26.9\%.
On Charades-FIG its median prediction is 8.0\,s, compared with 4.5\,s for \method on correctly retrieved videos (\cref{tab:supp_lengths}).

\noindent \textbf{TFVTG.}
TFVTG predicts a median duration of 8.7\,s on both Charades-FIG and DiDeMo-FIG.
Adding 2\,s per side lowers its VCMR R@1 at IoU 0.5 and 0.7 on both splits.
We also compare the two methods without a margin, giving each the annotated video and the original sentence.
At IoU 0.5, TFVTG leads on Charades-STA (50.0\% vs.\ 24.6\%), while \method leads on the second validation annotation of ActivityNet Captions (29.7\% vs.\ 27.0\%).

\noindent \textbf{Paired Comparison.}
We also compare the methods query by query using VideoSearch-R1's released predictions (\cref{tab:supp_paired}).
At IoU 0.5, the 2\,s margin gives \method a lead of 7.9 points on Charades-FIG and 2.2 points on DiDeMo-FIG.
At IoU 0.7, the DiDeMo-FIG result reverses and \method trails by 3.5 points.
Across the six dataset--threshold pairs, unpadded \method leads only on Charades-FIG at IoU 0.3.
\begin{table}[t]
\centering
\caption{\textbf{Paired Comparison with VideoSearch-R1}: \method\ minus the released predictions of \citet{lee2026videosearch} (points) on the same test queries.
VR R@1, and VCMR R@1 at IoU 0.3, 0.5 and 0.7 for the rows of \cref{tab:label_free}.}
\label{tab:supp_paired}
\vspace{-0.08in}
\small
\setlength{\tabcolsep}{5pt}
\begin{tabular}{lccc}
\toprule
& 0.3 & 0.5 & 0.7 \\
\midrule
\multicolumn{4}{l}{Charades-FIG} \\
VR R@1 & \multicolumn{3}{c}{$+8.2$} \\
\method & $+2.7$ & $-1.7$ & $-2.9$ \\
\quad w/ the whole activity & $+7.8$ & $+0.6$ & $-1.3$ \\
\quad w/ TFVTG's moment & $+6.7$ & $+3.7$ & $+0.3$ \\
\method\ w/ margin & $+10.1$ & $+7.9$ & $+6.5$ \\
\midrule
\multicolumn{4}{l}{DiDeMo-FIG} \\
VR R@1 & \multicolumn{3}{c}{$+6.0$} \\
\method & $-5.6$ & $-15.4$ & $-14.0$ \\
\quad w/ the whole activity & $-2.0$ & $-14.0$ & $-12.5$ \\
\quad w/ TFVTG's moment & $-3.2$ & $-10.4$ & $-12.3$ \\
\quad w/ DiDeMo's segments & $+11.6$ & $+11.0$ & $+9.9$ \\
\method\ w/ margin & $+10.6$ & $+2.2$ & $-3.5$ \\
\bottomrule
\end{tabular}
\vspace{-0.1in}
\end{table}

\subsection{Blind Baseline}
\label{sec:supp_blind}
Given only the sentence and video duration, gpt-6-luna reaches 0.8\% R@1 at IoU 0.5 on 3,720 Charades-STA test sentences (\cref{sec:supp_prompts}).
Nearly all its predictions cover at least 90\% of the video.
Supplying the dataset name and video identifier raises recall to 2.2\%.
On 2,000 sentences from the second validation annotation of ActivityNet Captions, where the annotated moments are longer, the blind baseline reaches 22.9\% at the same threshold.

\FloatBarrier
\section{Ablation Studies}
\label{sec:supp_ablations}

\subsection{Effect of the Margin}
\label{sec:supp_margin_effect}
\Cref{tab:test_margin} reports the test-set margin sweep at IoU 0.5 and 0.7 (\cref{fig:test_margin}).
On Charades-FIG, adding 2\,s to each side of VideoSearch-R1's intervals reduces recall at both thresholds.
The same margin improves the embedding-only baseline at IoU 0.5 on all three benchmarks.
\begin{table}[t]
\centering
\caption{\textbf{Margin around the Predicted Interval}: VCMR R@1 (\%) on the test splits at IoU 0.5 and 0.7 when the interval is widened by $w$ seconds on each side, or shrunk for $w<0$.
\method\ w/ margin uses $w=2$\,s.
VideoSearch-R1: the released predictions of \citet{lee2026videosearch} on Charades-FIG, scored against the annotation clipped to the video and widened the same way.
Embedding only: the half-height moment of the embedding-only baseline, at IoU 0.5.
Best per row in bold.}
\label{tab:test_margin}
\vspace{-0.08in}
\small
\setlength{\tabcolsep}{4pt}
\begin{tabular}{llccccccccccc}
\toprule
& IoU & \multicolumn{11}{c}{Margin $w$ (s)} \\
\cmidrule(lr){3-13}
& & $-5$ & $-3$ & $-2$ & $-1$ & $-0.5$ & 0 & $+0.5$ & $+1$ & $+2$ & $+3$ & $+5$ \\
\midrule
\multicolumn{13}{l}{\method} \\
Charades-FIG & 0.5 & 0.5 & 1.6 & 3.2 & 6.3 & 8.9 & 11.7 & 14.9 & 18.8 & 21.4 & \textbf{21.6} & 18.1 \\
 & 0.7 & 0.1 & 0.3 & 0.8 & 2.0 & 3.1 & 5.4 & 7.4 & 10.2 & \textbf{14.8} & 14.1 & 7.4 \\
DiDeMo-FIG & 0.5 & 0.5 & 1.3 & 2.9 & 6.8 & 9.7 & 14.9 & 20.1 & 26.5 & 32.4 & \textbf{32.5} & 18.9 \\
 & 0.7 & 0.1 & 0.3 & 0.8 & 2.1 & 4.0 & 5.9 & 8.6 & 11.2 & \textbf{16.4} & 14.9 & 8.5 \\
ActivityNet-FIG & 0.5 & 11.3 & 13.6 & 15.5 & 18.4 & 20.3 & 22.4 & 23.6 & 24.3 & 24.9 & 25.2 & \textbf{25.6} \\
 & 0.7 & 5.8 & 7.2 & 8.4 & 10.2 & 11.8 & 13.3 & \textbf{13.9} & \textbf{13.9} & \textbf{13.9} & 13.8 & 13.3 \\
\midrule
\multicolumn{13}{l}{VideoSearch-R1} \\
Charades-FIG & 0.5 & 0.3 & 2.7 & 7.7 & 12.2 & 13.2 & \textbf{13.5} & \textbf{13.5} & 13.3 & 12.3 & 10.9 & 8.0 \\
 & 0.7 & 0.0 & 0.6 & 2.1 & 6.2 & 7.7 & \textbf{8.3} & 8.1 & 7.4 & 5.3 & 4.2 & 2.4 \\
\midrule
\multicolumn{13}{l}{Embedding only} \\
Charades-FIG & 0.5 & 0.7 & 2.4 & 4.5 & 6.8 & 8.4 & 9.9 & 11.1 & 12.2 & \textbf{13.0} & 12.9 & 10.8 \\
DiDeMo-FIG & 0.5 & 1.4 & 2.9 & 4.4 & 7.3 & 9.6 & 11.8 & 13.7 & 15.3 & \textbf{16.6} & 15.4 & 10.8 \\
ActivityNet-FIG & 0.5 & 7.9 & 10.0 & 11.5 & 14.3 & 15.7 & 16.8 & 17.3 & 17.5 & 17.9 & 18.2 & \textbf{18.3} \\
\bottomrule
\end{tabular}
\vspace{-0.1in}
\end{table}

\noindent \textbf{Length of the Predicted Interval.}
Among queries for which \method ranks the annotated video first, 41--43\% of predicted intervals are shorter than half the reference duration (\cref{tab:supp_lengths}).
Even a perfectly centered interval of that length cannot reach IoU 0.5.
These short predictions account for 88--92\% of the misses that become hits with the 2\,s margin.
DiDeMo-FIG has a further source of error.
At least half its reference intervals span a single 5\,s segment, and 6.2\% of predictions on correctly retrieved videos start beyond the 30\,s annotation window.
\begin{table*}[t]
\centering
\caption{\textbf{Length of the Predicted Interval and of the Annotation} on the test splits, where the top-ranked video is the annotated one: quartiles of the length (s).
Short: the predicted interval is shorter than half of the annotated moment, which keeps its IoU below 0.5 (\%).
Centered: its center lies inside the annotated moment (\%).
Margin hits: the queries that reach IoU 0.5 only with the 2\,s margin, and the share of short intervals among them (\%).
After 30\,s: it starts after the 30\,s that DiDeMo annotates (\%).}
\label{tab:supp_lengths}
\vspace{-0.08in}
\setlength{\tabcolsep}{4pt}
\small
\begin{tabular}{lcccccccccccc}
\toprule
& & \multicolumn{3}{c}{Predicted interval (s)} & \multicolumn{3}{c}{Annotation (s)} & & & \multicolumn{2}{c}{Margin hits} & \\
\cmidrule(lr){3-5}\cmidrule(lr){6-8}\cmidrule(lr){11-12}
& Queries & 25\% & 50\% & 75\% & 25\% & 50\% & 75\% & Short & Centered & Queries & Short & After 30\,s \\
\midrule
Charades-FIG & 1,219 & 2.1 & 4.5 & 9.5 & 5.6 & 7.3 & 10.2 & 41.4 & 74.5 & 379 & 92.1 & -- \\
DiDeMo-FIG & 2,594 & 1.5 & 3.5 & 8.0 & 5.0 & 5.0 & 5.0 & 43.0 & 62.4 & 761 & 87.5 & 6.2 \\
ActivityNet-FIG & 11,189 & 4.2 & 12.5 & 41.4 & 11.0 & 26.8 & 55.9 & 40.7 & 68.8 & 753 & 87.8 & -- \\
\bottomrule
\end{tabular}
\vspace{-0.1in}
\end{table*}

\subsection{Component Analysis}
\label{sec:supp_components}
\noindent \textbf{Token Usage.}
Processing a query takes about 18.3k new input tokens on Charades-FIG and 39.3k on ActivityNet-FIG (\cref{tab:supp_cost}).
We amortize notes and the initial video turn across queries.
The localization row uses recorded token counts for 3,718 Charades-FIG, 4,013 DiDeMo-FIG and 16,242 ActivityNet-FIG query--video pairs, including retrieval misses.
For localization comparisons where a call is missing, we estimate input tokens from the frame count and the observed tokens per frame.
On Charades-FIG, a notes call averages 57 frames, 11.9k input tokens and 671 output tokens.
A text-only query turn adds about 0.5k new input tokens.
A second localization pass is needed for 0.1\% of Charades-FIG queries and 77\% of ActivityNet-FIG queries.
\begin{table}[t]
\centering
\caption{\textbf{Mean tokens per query on the test splits.} We share the cost of a video's notes and initial localization turn across queries. For localization, New and Cached are the uncached and cached input counts. For notes and ranking, New includes all input tokens, of which fewer than 0.05k per query were cached. Output counts generated tokens.}
\label{tab:supp_cost}
\vspace{-0.08in}
\small
\setlength{\tabcolsep}{4pt}
\begin{tabular}{lccccccccc}
\toprule
& \multicolumn{3}{c}{Charades-FIG} & \multicolumn{3}{c}{DiDeMo-FIG} & \multicolumn{3}{c}{ActivityNet-FIG} \\
\cmidrule(lr){2-4}\cmidrule(lr){5-7}\cmidrule(lr){8-10}
& New & Cached & Output & New & Cached & Output & New & Cached & Output \\
\midrule
Notes (shared) & 4.3k & 0.0k & 241 & 4.1k & 0.0k & 186 & 4.6k & 0.0k & 324 \\
Ranking call & 6.2k & 0.0k & 33 & 7.5k & 0.0k & 33 & 11.0k & 0.0k & 34 \\
Localization & 7.8k & 21.2k & 29 & 11.2k & 30.5k & 34 & 23.6k & 29.7k & 45 \\
\midrule
Total & 18.3k & 21.2k & 302 & 22.8k & 30.5k & 254 & 39.3k & 29.7k & 404 \\
\bottomrule
\end{tabular}
\vspace{-0.1in}
\end{table}

\noindent \textbf{Video Prompt Caching.}
On Charades-FIG, caching reduces new localization input from 22.8k to 7.8k tokens per query and raises VCMR R@1 at IoU 0.5 by 0.9 points (\cref{tab:supp_threads}).
On DiDeMo-FIG, new input falls from 34.6k to 11.2k and recall rises by 1.8 points.
ActivityNet-FIG sees the smallest proportional saving (46.8k to 23.6k), and recall at IoU 0.5 stays at 24.9\%.
Its frequent second pass uses a separate call.
Keeping prior queries and answers in the conversation makes predictions sensitive to query order.
On Charades-FIG development queries, recall at IoU 0.5 was 68.4\% in the original order and 64.5\% in reverse order, compared with 66.3\% when we discarded each turn.
On test queries, keeping the history brought recall back to the one-call-per-query result (20.5\%).
\begin{table*}[t]
\centering
\caption{\textbf{Video Prompt Caching against One Localization Call per Query} (\%, the interval widened by 2\,s).
Dev: R@1 in the annotated video on the 487 queries of the Charades-FIG development set (210 videos); test: VCMR R@1 in \method's top-ranked video.
Tokens: localization input tokens per query, all and not read from the prompt cache (new); the first turn of a video's conversation is shared over its queries.
Difference: against one call per query at IoU 0.5 / 0.7.}
\label{tab:supp_threads}
\vspace{-0.08in}
\setlength{\tabcolsep}{4pt}
\small
\begin{tabular}{llccccc}
\toprule
& & \multicolumn{2}{c}{R@1} & \multicolumn{2}{c}{Tokens} & \\
\cmidrule(lr){3-4}\cmidrule(lr){5-6}
& & 0.5 & 0.7 & All & New & Difference 0.5 / 0.7 \\
\midrule
Charades-FIG dev & one call per query & 61.8 & 40.5 & 23.4k & 20.5k &  \\
 & video prompt caching & 66.3 & 43.1 & 33.2k & 10.1k & +4.5 / +2.7 \\
 & video prompt caching, every turn kept & 68.4 & 44.1 & 33.2k & 9.4k & +6.6 / +3.7 \\
 & same, queries in reversed order & 64.5 & 41.9 & 33.3k & 8.9k & +2.7 / +1.4 \\
\midrule
Charades-FIG test & one call per query & 20.5 & 14.1 & 22.8k & 22.8k &  \\
 & video prompt caching & 21.4 & 14.8 & 29.0k & 7.8k & +0.9 / +0.7 \\
 & video prompt caching, every turn kept & 20.5 & 14.3 & 29.1k & 7.6k & +0.0 / +0.2 \\
\midrule
DiDeMo-FIG test & one call per query & 30.7 & 15.4 & 34.6k & 34.6k &  \\
 & video prompt caching & 32.4 & 16.4 & 41.7k & 11.2k & +1.8 / +1.0 \\
\midrule
ActivityNet-FIG test & one call per query & 24.9 & 14.2 & 46.8k & 46.8k &  \\
 & video prompt caching & 24.9 & 13.9 & 53.3k & 23.6k & +0.0 / $-$0.3 \\
\bottomrule
\end{tabular}
\vspace{-0.1in}
\end{table*}

\noindent \textbf{Ranking Variants.}
Adding notes-based reranking and fusion to query-bank search raises development VR R@1 by 5.4 points on Charades-FIG, 0.6 on DiDeMo-FIG and 4.6 on ActivityNet-FIG (\cref{tab:supp_dev}).
The query bank itself contributes 4.3, 4.0 and 0.6 points, respectively.
\begin{table*}[t]
\centering
\caption{\textbf{Ranking Variants on the Development Sets} (487 queries on Charades-FIG, 500 on DiDeMo-FIG and 500 on ActivityNet-FIG; R@1 in \%, the interval widened by 2\,s).
Tokens: input tokens per query of the ranking stage and the localization, with one localization call per query; the notes of a video are shared over all queries of its split.}
\label{tab:supp_dev}
\vspace{-0.08in}
\setlength{\tabcolsep}{3pt}
\small
\begin{tabular}{lccccccccc}
\toprule
& \multicolumn{3}{c}{Charades-FIG} & \multicolumn{3}{c}{DiDeMo-FIG} & \multicolumn{3}{c}{ActivityNet-FIG} \\
\cmidrule(lr){2-4}\cmidrule(lr){5-7}\cmidrule(lr){8-10}
& VR & VCMR 0.5 & Tokens & VR & VCMR 0.5 & Tokens & VR & VCMR 0.5 & Tokens \\
\midrule
Search & 24.0 & 11.5 & 0.0k & 68.6 & 20.2 & 0.0k & 61.6 & 21.2 & 0.0k \\
Search with query bank & 28.3 & 14.2 & 0.0k & 72.6 & 21.4 & 0.0k & 62.2 & 21.2 & 0.0k \\
\method\ w/o query bank & 25.9 & 15.2 & 35.5k & 68.8 & 33.2 & 46.1k & 66.6 & 25.8 & 63.6k \\
\method\ w/o fusion & 32.4 & 19.3 & 35.2k & 66.8 & 31.8 & 45.6k & 64.2 & 26.8 & 62.7k \\
\method & \cellcolor{LightCyan}33.7 & \cellcolor{LightCyan}19.3 & \cellcolor{LightCyan}35.3k & \cellcolor{LightCyan}73.2 & \cellcolor{LightCyan}35.2 & \cellcolor{LightCyan}45.7k & \cellcolor{LightCyan}66.8 & \cellcolor{LightCyan}26.6 & \cellcolor{LightCyan}62.7k \\
\bottomrule
\end{tabular}
\vspace{-0.1in}
\end{table*}

On Charades-FIG, ranking by notes alone, without the query bank or fusion, gives 30.6\% VR R@1.
Expanding the list to 30 or 50 candidates gives 25.1\% and 24.8\%.
Reranking the top three with 16 or 40 frames per video gives 30.2\% and 29.2\%.
A prompt that first asks for the closest matching step and any contradictory details gives 32.4\%.
The full method reaches 33.7\%.
Using notes alone for localization matches frame-based localization on Charades-FIG (19.3\% VCMR R@1 at IoU 0.5 with the margin).
It loses 6.8 points on DiDeMo-FIG and 1.6 on ActivityNet-FIG.
The 64-frame notes budget provides sparser coverage of longer videos.

\noindent \textbf{Beyond the Top-Ranked Video.}
We ask the VLM to check the top three videos in order and localize in the first one it accepts (\cref{sec:supp_prompts}).
If it rejects all three, we use the original top-ranked video.
This raises VR R@1 by 3.3--4.1 points across the test splits and improves VCMR R@1 at IoU 0.5 on each (\cref{tab:supp_rounds}).
On Charades-FIG, input rises from 18.3k to 37.6k tokens per query.
The VLM rejects 41\% of incorrect first choices and 10\% of correct ones.
Its 4.1-point VR gain recovers about a quarter of the 15.9 points available if an oracle always selects the annotated video when it appears in the top three (32.8\% to 48.7\%).
The verification call also predicts an interval.
Using that interval in the selected video, with the margin, gives lower recall than a separate localization call on Charades-FIG and DiDeMo-FIG and higher recall on ActivityNet-FIG.
\begin{table}[t]
\centering
\caption{\textbf{Beyond the Top-Ranked Video} on the test splits (\%): VR R@1, and VCMR R@1 at IoU 0.5 without and with the 2\,s margin.
W/ fallback to the next videos: the VLM checks the top three videos of the final ranking in order, and \method\ predicts the interval in the first one that it accepts, or in the top-ranked video if it accepts none.
The interval of the fallback call: in the same video, the interval that the VLM returns with its check.
Upper bound: the annotated video is ranked first whenever it is among the top 3 or the top 10 of the final ranking, and \method\ predicts the interval in it.}
\label{tab:supp_rounds}
\vspace{-0.08in}
\small
\setlength{\tabcolsep}{4pt}
\begin{tabular}{lccccccccc}
\toprule
& \multicolumn{3}{c}{Charades-FIG} & \multicolumn{3}{c}{DiDeMo-FIG} & \multicolumn{3}{c}{ActivityNet-FIG} \\
\cmidrule(lr){2-4}\cmidrule(lr){5-7}\cmidrule(lr){8-10}
& VR & 0.5 & +2\,s & VR & 0.5 & +2\,s & VR & 0.5 & +2\,s \\
\midrule
\method & 32.77 & 11.75 & 21.37 & 64.64 & 14.88 & 32.44 & 65.96 & 22.39 & 24.91 \\
\quad w/ fallback to the next videos & 36.83 & 12.90 & 24.46 & 68.08 & 15.67 & 34.59 & 69.23 & 23.45 & 26.26 \\
\quad w/ fallback, the interval of the fallback call & 36.83 & 13.20 & 24.17 & 68.08 & 15.42 & 31.62 & 69.23 & 24.85 & 27.08 \\
\quad upper bound, annotated video in the top 3 & 48.71 & 17.18 & 32.26 & 81.78 & 18.81 & 40.74 & 79.97 & 26.72 & 29.78 \\
\quad upper bound, annotated video in the top 10 & 62.42 & 21.56 & 41.45 & 89.86 & 20.93 & 44.65 & -- & -- & -- \\
\bottomrule
\end{tabular}
\vspace{-0.1in}
\end{table}

\subsection{Source of the Query Bank}
\label{sec:supp_bank}
\Cref{tab:test_bank} evaluates the full pipeline with training queries in the bank.
Here we isolate the search stage and vary the bank's source and size, keeping $K_b$ and $\alpha$ fixed (\cref{tab:supp_bank}).
With 1,000--10,000 training queries, VR R@1 is within 0.7 points of using the full training bank on each benchmark.
Small test batches behave differently across datasets.
A bank of 20 test queries improves DiDeMo-FIG by 2.34 points over search alone, while ActivityNet-FIG stays below search alone for batch sizes up to 500.
Removing each query from its own full-split bank changes VR R@1 by at most 0.06 points.
\begin{table}[t]
\centering
\caption{\textbf{Queries in the Query Bank}: VR (\%) of the search on the test splits with other queries as the query bank $\mathcal{Q}$ ($K_b=20$, $\alpha=0.75$).
The split's queries: all queries of the evaluated split, as in \method.
Training queries: the queries of the benchmark's training split (12,408, 32,903, 37,261), all or a random subset.
Batches: the split is cut at random into batches, and each batch is its own $\mathcal{Q}$.
Stream: the queries arrive in a random order, and $\mathcal{Q}$ holds those that have arrived.
$K_b$ grown with the bank: $K_b$ is proportional to the size of the bank, 20 at 2,693 queries.
Starting from the training queries: they are in the bank before the first arrival.
Rows with random draws average 5 draws.}
\label{tab:supp_bank}
\vspace{-0.08in}
\small
\setlength{\tabcolsep}{4pt}
\begin{tabular}{lcccccc}
\toprule
& \multicolumn{2}{c}{Charades-FIG} & \multicolumn{2}{c}{DiDeMo-FIG} & \multicolumn{2}{c}{ActivityNet-FIG} \\
\cmidrule(lr){2-3}\cmidrule(lr){4-5}\cmidrule(lr){6-7}
$\mathcal{Q}$ & R@1 & R@10 & R@1 & R@10 & R@1 & R@10 \\
\midrule
No query bank & 25.97 & 58.52 & 57.06 & 86.99 & 60.11 & 86.84 \\
The split's queries (\method) & 28.36 & 62.42 & 61.48 & 89.86 & 62.28 & 87.60 \\
\quad without the query itself & 28.41 & 62.39 & 61.53 & 89.78 & 62.22 & 87.59 \\
\midrule
All training queries & 28.04 & 61.91 & 59.61 & 89.04 & 60.37 & 86.92 \\
\quad 100 training queries & 26.33 & 59.97 & 59.62 & 89.43 & 58.69 & 86.27 \\
\quad 500 training queries & 27.36 & 61.47 & 60.00 & 89.59 & 59.60 & 86.75 \\
\quad 1,000 training queries & 27.52 & 61.52 & 60.19 & 89.45 & 59.97 & 87.06 \\
\quad 2,000 training queries & 27.61 & 61.76 & 60.25 & 89.40 & 60.44 & 87.16 \\
\quad 5,000 training queries & 27.99 & 62.03 & 60.16 & 89.24 & 60.80 & 87.23 \\
\quad 10,000 training queries & 27.94 & 62.03 & 60.14 & 89.19 & 60.62 & 87.14 \\
\midrule
Batches of 20 queries & 26.16 & 59.48 & 59.40 & 89.37 & 57.48 & 85.79 \\
Batches of 50 queries & 26.22 & 59.62 & 59.64 & 89.37 & 58.35 & 86.11 \\
Batches of 100 queries & 26.40 & 60.12 & 59.70 & 89.53 & 58.71 & 86.22 \\
Batches of 200 queries & 27.09 & 60.77 & 60.23 & 89.64 & 59.10 & 86.44 \\
Batches of 500 queries & 27.49 & 61.45 & 60.26 & 89.70 & 59.64 & 86.80 \\
Batches of 1,000 queries & 27.90 & 61.80 & 60.62 & 89.75 & 60.24 & 87.07 \\
Batches of 2,000 queries & 28.25 & 62.27 & 60.81 & 89.77 & 60.62 & 87.23 \\
Stream & 27.91 & 62.02 & 61.04 & 89.71 & 61.46 & 87.37 \\
\quad without the query itself & 27.84 & 61.97 & 60.99 & 89.75 & 61.40 & 87.37 \\
\quad $K_b$ grown with the bank & 27.85 & 61.62 & 60.29 & 89.27 & 61.42 & 87.28 \\
\quad starting from the training queries & 28.12 & 62.08 & 60.00 & 89.35 & 60.97 & 87.24 \\
\bottomrule
\end{tabular}
\vspace{-0.1in}
\end{table}

\FloatBarrier
\section{Streaming Queries}
\label{sec:supp_stream}
\noindent \textbf{End-to-End Evaluation.}
For \cref{tab:test_stream}, we replay one random arrival order per test split (seed 0).
We score each query when it arrives.
To compare with the batch run, we reuse notes for videos already described and rerank whenever the ordered list of ten candidates differs from the batch list.
After ranking, we reuse the prediction for an unchanged query--video pair.
If the stream newly ranks the annotated video first, we obtain a localization prediction for that pair.
A query that retrieves the wrong video already counts as a VCMR miss regardless of its interval.

\noindent \textbf{Query Bank.}
At arrival $n$, the bank contains the first $n$ queries, including the current one.
For each video, we average its $\min(20,n)$ highest scores and apply the same penalty weight as in batch search.
To estimate the gain at a given $n$ in \cref{fig:stream}, we draw random subsets of size $n$ until we have scored at least 100,000 query instances.
At the full split size we score the split once.
\Cref{tab:supp_stream} instead groups arrival positions across five random orders.
On ActivityNet-FIG, the bank reduces search VR R@1 by 1.90 points for arrivals 1--100, by 0.89 for 101--500 and by 0.54 for 501--1,000.
Later bins show a gain.
Excluding the current query leaves the early losses in place.
Initializing the bank with training queries removes those losses on ActivityNet-FIG but lowers recall by 2.20 points for the first 100 Charades-FIG arrivals relative to search alone.
\begin{table}[t]
\centering
\caption{\textbf{Query Bank with Streaming Queries}: VR R@1 of the search with the query bank minus that without a query bank (points), by arrival position, averaged over 5 random orders; the last row gives VR R@1 over all queries.
Stream: $\mathcal{Q}$ holds the queries that have arrived, the query itself included; w/o itself: only the earlier ones; from training: $\mathcal{Q}$ also holds the benchmark's training queries from the start.}
\label{tab:supp_stream}
\vspace{-0.08in}
\small
\setlength{\tabcolsep}{4pt}
\begin{tabular}{lccccccccc}
\toprule
& \multicolumn{3}{c}{Charades-FIG} & \multicolumn{3}{c}{DiDeMo-FIG} & \multicolumn{3}{c}{ActivityNet-FIG} \\
\cmidrule(lr){2-4}\cmidrule(lr){5-7}\cmidrule(lr){8-10}
Arrival & Stream & w/o itself & from training & Stream & w/o itself & from training & Stream & w/o itself & from training \\
\midrule
1--100 & $-$0.60 & $-$0.40 & $-$2.20 & +0.80 & +0.40 & +1.60 & $-$1.90 & $-$1.90 & +0.41 \\
101--500 & +1.75 & +1.65 & +2.90 & +4.70 & +4.60 & +2.90 & $-$0.89 & $-$1.04 & +0.36 \\
501--1,000 & +1.44 & +1.36 & +1.60 & +3.16 & +2.96 & +2.68 & $-$0.54 & $-$0.75 & +0.17 \\
1,001--2,000 & +2.14 & +1.98 & +2.52 & +3.36 & +3.26 & +2.36 & +0.38 & +0.21 & +0.58 \\
2,001 on & +2.18 & +2.14 & +2.18 & +4.50 & +4.54 & +3.36 & +1.63 & +1.58 & +0.95 \\
\midrule
VR R@1 (\%) & 27.91 & 27.84 & 28.12 & 61.04 & 60.99 & 60.00 & 61.46 & 61.40 & 60.97 \\
\bottomrule
\end{tabular}
\vspace{-0.1in}
\end{table}

\noindent \textbf{Video Prompt Caching.}
We estimate streaming token use by replaying five random orders with a cache that retains a video's frames for a fixed number of arrivals after their last use (\cref{fig:stream}, right).
A query incurs the cost of the initial video turn again if its frames have expired.
The median gap between successive queries retrieving the same video is 481 arrivals on Charades-FIG and 2,399 on ActivityNet-FIG.
In this simulation, ActivityNet-FIG needs about five times the retention period to recover the same share of batch savings (\cref{sec:stream}).

\FloatBarrier
\section{Annotation Analysis and Examples}
\label{sec:supp_analysis}

\subsection{Human Annotations}
\label{sec:supp_annotators}
\begin{figure*}[h]
\centering
\includegraphics[width=\linewidth]{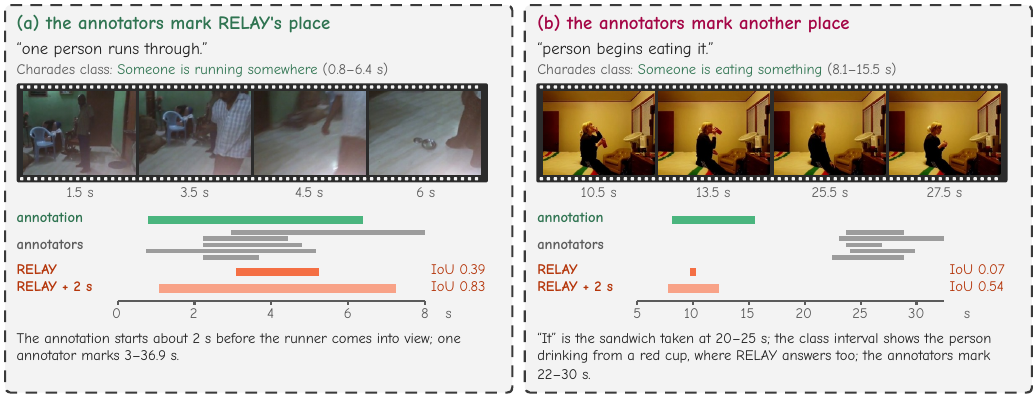}
\vspace{-0.2in}
\caption{\textbf{When a margin helps and when it hides an error.} Each panel shows the Charades activity-class interval, independent annotators' intervals and the model prediction for one Charades-STA test sentence.
The displayed IoUs use the activity-class interval as reference. (a) The runner appears after the class interval begins. (b) The class interval covers drinking from a cup.
The annotators place the queried action, eating a sandwich, later in the video.}
\label{fig:margin_examples}
\end{figure*}
On Charades-STA, widening each human annotator's interval by 2\,s gives the highest mean R@1 against the benchmark among the margins tested (73.7\% at IoU 0.5, \cref{tab:dev_margin}).
For 105 sentences, the model already reaches IoU 0.5 against the annotators' median interval but needs the margin to reach that threshold against the benchmark.
The running example in \cref{fig:margin_examples}(a) shows this difference.
For 12 other sentences, widening produces a benchmark hit even though the original prediction has no overlap with the annotators' median interval.
In panel (b), the benchmark IoU rises from 0.07 to 0.54 while the model still places the action at the drinking scene.

\noindent \textbf{Refined Annotations.}
TimeLens-Bench~\cite{zhang2026timelens} revises temporal boundaries and rewrites some sentences.
To compare boundaries for the same wording, we use 872 Charades-STA and 1,314 ActivityNet Captions sentences retained verbatim (\cref{tab:supp_refined}).
On Charades-STA, adding the margin raises R@1 at IoU 0.5 from 26.4\% to 74.5\% against the original boundaries.
Against the revised boundaries, recall changes from 41.7\% to 41.6\%.
On ActivityNet Captions, the corresponding changes are 36.4\% to 41.6\% and 52.7\% to 51.4\%.
At IoU 0.5 and 0.7, the unpadded predictions agree more often with the revised boundaries than the original annotations themselves do.
The Charades-FIG rows examine queries whose original sentence was retained.
Each method is scored on its own correctly retrieved videos, so the VideoSearch-R1 and \method rows cover different query sets.
Within their respective subsets, both TFVTG and VideoSearch-R1 score lower against the revised boundaries than against the original ones at all three thresholds.
\begin{table}[t]
\centering
\caption{\textbf{Refined Annotations of TimeLens-Bench}~\cite{zhang2026timelens}: R@1 (\%) against the original annotation and against the refined one, on the sentences that TimeLens-Bench keeps verbatim.
Annotation: the original annotation, scored against the refined one.
Charades-FIG: the test queries that rewrite a kept sentence, where the top-ranked video of \method\ or of VideoSearch-R1 is the annotated one.}
\label{tab:supp_refined}
\vspace{-0.08in}
\small
\setlength{\tabcolsep}{4pt}
\begin{tabular}{lcccccc}
\toprule
& \multicolumn{3}{c}{vs.\ annotation} & \multicolumn{3}{c}{vs.\ refined annotation} \\
\cmidrule(lr){2-4}\cmidrule(lr){5-7}
Moment & 0.3 & 0.5 & 0.7 & 0.3 & 0.5 & 0.7 \\
\midrule
\multicolumn{7}{l}{Charades-STA (872 sentences), in the annotated video} \\
Annotation & -- & -- & -- & 65.9 & 36.1 & 14.6 \\
\method & 48.3 & 26.4 & 12.2 & 62.5 & 41.7 & 22.8 \\
\method\ w/ margin & 85.9 & 74.5 & 53.0 & 70.9 & 41.6 & 18.7 \\
\midrule
\multicolumn{7}{l}{ActivityNet Captions (1,314 sentences), in the annotated video} \\
Annotation & -- & -- & -- & 65.1 & 50.2 & 35.9 \\
\method & 50.9 & 36.4 & 24.7 & 63.2 & 52.7 & 40.6 \\
\method\ w/ margin & 58.4 & 41.6 & 26.0 & 66.7 & 51.4 & 35.5 \\
\midrule
\multicolumn{7}{l}{Charades-FIG (296 queries), in \method's top-ranked video} \\
Annotation & -- & -- & -- & 63.2 & 33.4 & 15.5 \\
\method & 61.8 & 37.8 & 18.2 & 64.9 & 46.3 & 27.4 \\
\quad w/ the whole activity & 78.4 & 51.0 & 24.3 & 66.2 & 45.6 & 27.4 \\
\quad w/ TFVTG's moment & 74.7 & 55.7 & 28.7 & 47.6 & 26.7 & 13.2 \\
\midrule
\multicolumn{7}{l}{Charades-FIG (196 queries), in VideoSearch-R1's top-ranked video} \\
VideoSearch-R1 & 79.1 & 63.3 & 35.7 & 55.6 & 35.7 & 15.3 \\
\bottomrule
\end{tabular}
\vspace{-0.1in}
\end{table}

\subsection{Qualitative Results}
\label{sec:supp_examples}
\begin{figure*}[h]
\centering
\includegraphics[width=\linewidth]{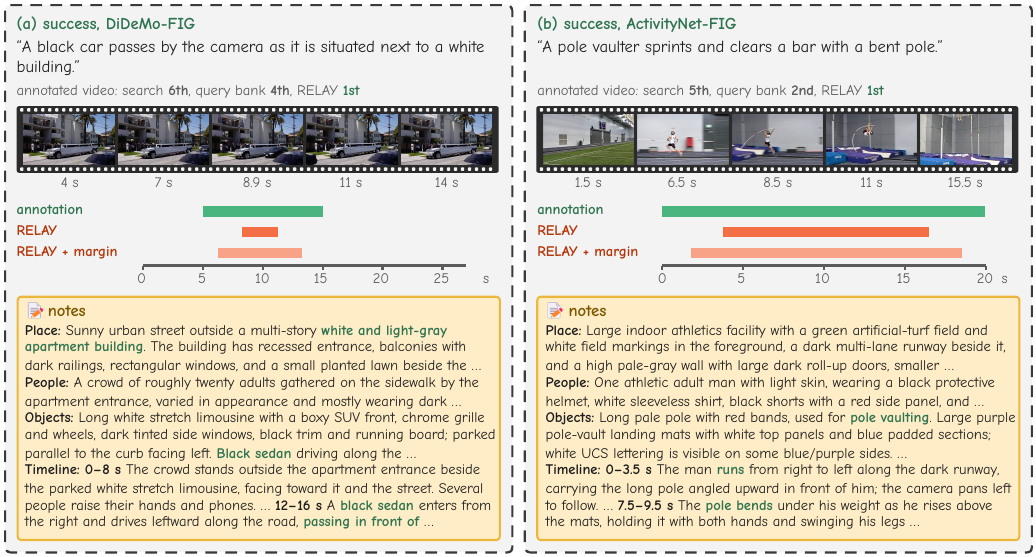}
\vspace{-0.2in}
\caption{\textbf{Retrieval and localization on two test queries.} (a) The DiDeMo-FIG reference covers two consecutive 5\,s segments as a black car passes the camera. (b) The ActivityNet-FIG reference covers the entire pole-vault video.}
\label{fig:examples_more}
\end{figure*}

In the DiDeMo-FIG example, the notes distinguish the moving black sedan from the parked white limousine.
In the ActivityNet-FIG example, the timeline records when the vaulter's pole bends.
The annotated videos move from sixth to fourth and from fifth to second after query-bank normalization.
Both rank first after reranking (\cref{fig:examples_more}).

\section{Limitations}
\label{sec:supp_limitations}
On Charades-FIG, search leaves the annotated video outside the top ten for 37.6\% of test queries.
Reranking cannot recover these misses.
Even when search retrieves the right video, 64 uniformly sampled frames can miss a brief action or a small object.
Choosing a margin requires labeled examples, and matching the benchmark boundaries more closely can mask a localization error (\cref{fig:margin_examples}).

Batch processing lets us estimate the query-bank penalty from all queries and share video frames across repeated requests.
In a stream, a small initial bank can reduce recall, and long gaps between requests for a video reduce cache reuse (\cref{fig:stream}).
The main experiments use a proprietary model.
The open-model comparison covers only Charades-FIG (\cref{sec:supp_open}).

\section{Prompts}
\label{sec:supp_prompts}
The following prompts cover \method, its variants and the blind baseline.
In visual prompts, each timestamp in seconds precedes its frame.

\subsection{\method}
Ranking makes two calls (\cref{sec:supp_rank}), and localization one call per run (\cref{sec:supp_localize}).
The first run shares turn 1 among the queries of a video (video prompt caching).
The second run sends the frames of its window and turn 2 in one message.

\begin{promptbox}[Video Notes\hfill\mdseries VLM, once per candidate video]{promptgray}
\promptrole{System}
You watch a video and write notes from which someone who cannot see it can later tell whether a description of a moment fits the video, and when that moment happens. You see frames of the whole video in time order, each after its time in seconds. Write:\par
\promptitem - place: the kind of room or place and what is in it (furniture, fixtures, colors);\par
\promptitem - people: every person, with what they look like and what they wear (colors, clothing);\par
\promptitem - objects: every object that a person touches, holds or uses, and other objects that stand out, each with its color, material and other visible details;\par
\promptitem - timeline: everything that happens, in time order, each step with its start and end in seconds: who does what, to which object, where, and in which direction (opening or closing, picking up or putting down, sitting down or standing up). Include small actions, such as looking at something, smiling, sneezing or drinking.\par
Write only what the frames show.\par
\promptrole{User}
\promptnote{the frames of the whole video, each after its time in seconds}\par
The frames are \promptfield{step} s apart; the video lasts \promptfield{duration} s. Write the notes.\par
\end{promptbox}

\begin{promptbox}[Notes-Based Ranking\hfill\mdseries LLM, text only, once per query]{promptgray}
\promptrole{System}
You rank videos by how well they show the moment that a description describes. You cannot see the videos: you see numbered videos, each as notes that someone wrote while watching it (the place, the people, the objects, and a timeline of what happens with times in seconds), and then the description. Order all the videos by their numbers, from the one that best shows the described moment to the one that shows it least. Check the details: the action and its direction (opening or closing, sitting down or standing up), the objects, the people and what they wear, the place. Notes can leave a detail out: a detail that the notes do not mention counts less against a video than a detail that they contradict.\par
\promptrole{User}
Video \promptfield{number}:\par
\promptfield{notes}\par
\promptnote{repeated for each candidate, in the order of the embedding-based ranking}\par
\smallskip
Description: \promptfield{query}\par
Order the \promptfield{count} videos, best first.\par
\end{promptbox}

\begin{promptbox}[Localization\hfill\mdseries VLM, video prompt caching]{promptgray}
\promptrole{System}
You find the moment of a video that a description describes. You see frames of a part of the video in time order, each after its time in seconds, and then the description and what to answer. Answer times in seconds, as precisely as the spacing of the frames allows.\par
\promptrole{User, turn 1 (once per video)}
\promptnote{the frames of the whole video, each after its time in seconds}\par
Descriptions of moments of this video follow, one per message, each with what to answer. First answer the start and the end of the whole video.\par
\promptrole{Model}
\promptnote{the start and the end of the whole video}\par
\promptrole{User, turn 2 (once per query)}
The frames are \promptfield{step} s apart.\par
Description: \promptfield{query}\par
Answer the start and the end of the moment.\par
\end{promptbox}

\subsection{Variants}
Each variant replaces one call of \method.
Without video prompt caching, each localization call sends the frames and the text of turn 2 in one message.
Fallback to the next videos and two intervals change only turn 1 or the last line of turn 2 of localization.
With subtitles (TVR), the notes and the localization read the subtitle lines after the frames.

\begin{promptbox}[Frames for Ranking\hfill\mdseries VLM, in place of notes-based ranking]{promptgray}
\promptrole{System}
You rank videos by how well they show the moment that a description describes. You see numbered videos, each as frames from its whole length in time order, each frame after its time in seconds, and then the description. Order all the videos by their numbers, from the one that best shows the described moment to the one that shows it least. Check the details: the action and its direction (opening or closing, sitting down or standing up), the objects, the people and what they wear, the place.\par
\promptrole{User}
Video \promptfield{number}:\par
\promptnote{the frames of the whole video, each after its time in seconds}\par
\promptnote{repeated for each candidate, in the order of the embedding-based ranking}\par
\smallskip
Description: \promptfield{query}\par
Order the \promptfield{count} videos, best first.\par
\end{promptbox}

\begin{promptbox}[Localization from the Notes\hfill\mdseries LLM, text only, in place of localization]{promptgray}
\promptrole{System}
You find the moment of a video that a description describes. You cannot see the video: you see notes that someone wrote while watching it (the place, the people, the objects, and a timeline of what happens with times in seconds), and then the description. Answer the start and the end of the moment in seconds.\par
\promptrole{User}
The video lasts \promptfield{duration} s.\par
Notes:\par
\promptfield{notes}\par
Description: \promptfield{query}\par
\end{promptbox}

\begin{promptbox}[Fallback to the Next Videos and Two Intervals\hfill\mdseries changes to localization]{promptgray}
\promptrole{Fallback to the next videos: last line of turn 2, in each of the top three videos}
Answer whether this video shows the moment, and the start and the end of the moment. Answer that it is not shown only if the frames show that the description does not fit this video; then answer the start and the end of the part of the video that comes closest to it.\par
\promptrole{Two intervals: turn 1}
Descriptions of moments of this video follow, one per message, each with what to answer. First answer the start and the end of the whole video as both intervals.\par
\promptrole{Two intervals: last line of turn 2}
Answer two intervals: the action that the description names, from when it begins until it ends, and the whole activity around it, from when getting ready for the action begins (reaching for or walking up to what it involves) until the action is finished.\par
\end{promptbox}

\begin{promptbox}[W/ Subtitles\hfill\mdseries TVR, changes to notes and localization]{promptgray}
\promptrole{Notes: last line of the system prompt}
Write only what the frames and the subtitles show; where the subtitles tell who a person is, name that person.\par
\promptrole{Notes and localization: after the frames, in the same message}
Subtitles, each line with its start and end in seconds and, where known, the speaker before the colon:\par
\promptfield{start}-\promptfield{end} s: \promptfield{text}\par
\promptnote{one such line for each subtitle that overlaps the window of the frames}\par
\end{promptbox}

\subsection{Blind Baseline}
The blind baseline of \cref{sec:supp_blind} sees no frames.

\begin{promptbox}[Blind Baseline\hfill\mdseries LLM, text only]{promptgray}
\promptrole{System}
You find the moment of a video that a description describes. You cannot see the video: you see how long it lasts and the description, sometimes also the dataset that the video comes from and its identifier there. Answer the start and the end of the moment in seconds.\par
\promptrole{User}
The video lasts \promptfield{duration} s.\par
Description: \promptfield{query}\par
\promptrole{User, with the dataset and the video\textquotesingle{}s identifier}
The video is \promptfield{video} of \promptfield{dataset} and lasts \promptfield{duration} s.\par
Description: \promptfield{query}\par
\end{promptbox}

\end{document}